\documentclass[11pt]{article}

\usepackage[final]{acl}

\usepackage{amsmath}
\usepackage{amsfonts}
\usepackage{booktabs}
\usepackage{subcaption}
\usepackage{mdframed}
\usepackage{placeins}
\usepackage{multirow}
\usepackage[table]{xcolor}
\definecolor{myteal}{HTML}{A4DDD3} 

\usepackage{times}
\usepackage{latexsym}

\usepackage[T1]{fontenc}

\usepackage[utf8]{inputenc}

\usepackage{microtype}

\usepackage{inconsolata}

\usepackage{graphicx}

\title{Evidence-Aligned Entity Verification for Hallucination Detection in Retrieval-Augmented Generation}

\author{
  Runsong Jia\textsuperscript{1} \quad
  Zhen Fang\textsuperscript{1}\thanks{\ Corresponding authors.} \quad
  Mengjia Wu\textsuperscript{1} \quad
  Jie Lu\textsuperscript{1} \quad
  Yi Zhang\textsuperscript{1}\footnotemark[\value{footnote}] \\
  \textsuperscript{1}University of Technology Sydney, Sydney, Australia \\
  \texttt{runsong.jia@student.uts.edu.au} \\
  \texttt{\{zhen.fang, mengjia.wu, jie.lu, yi.zhang\}@uts.edu.au}
}

\begin{document}
\maketitle
\begin{abstract}
Hallucination detection is crucial for large language models (LLMs), as hallucinated content creates significant barriers in applications requiring factual accuracy. Current detection methods mainly depend on internal signals like uncertainty and self-consistency checks, using the model's pre-trained knowledge to identify unreliable outputs. However, pre-trained knowledge may become outdated and has coverage limitations, especially for specialized or recent information. To address these limitations, retrieval-augmented generation (RAG) has emerged as a promising solution by retrieving relevant evidence at inference time, grounding outputs beyond the model’s parametric knowledge. In this paper, we target a critical and practical learning problem \textit{RAG-based hallucination detection} (RHD), where RAG is employed to enhance hallucination detection by addressing information updating challenges. To address RHD, we propose a novel method \textit{Evidence-Aligned Entity Verification} (EAEV), which detects entity-level hallucinations by leveraging RAG to align generated entities with retrieved evidence contexts. Specifically, EAEV evaluates entity-evidence alignment through three complementary dimensions and introduces counterfactual stability analysis to ensure robust alignments under evidence perturbations. Experiments across multiple RAG benchmarks demonstrate that EAEV achieves consistent improvements over existing methods with strong generalization capabilities.
\end{abstract}

\section{Introduction}
\label{sec:intro}

The deployment of large language models (LLMs) in practical applications faces a critical challenge: models frequently generate factually incorrect or inconsistent content, known as hallucinations \citep{ji2023survey}. This problem poses significant risks in domains where accuracy is essential, such as medical diagnosis, educational assistance, and financial advisory services \citep{tang2024medirag,li2023halueval}. As organizations increasingly rely on LLMs for complex tasks, the consequences of undetected hallucinations can range from misinformation propagation to decision-making failures, making robust hallucination detection an urgent priority for trustworthy AI deployment.

Existing hallucination detection methods have established foundations across diverse paradigms, including uncertainty and consistency based detectors that leverage internal signals~\citep{manakul2023selfcheckgpt,farquhar2024semantic,li2023halueval,zhang2023enhancing}, as well as evidence based verification using NLI models or LLM judges to check consistency against provided text~\citep{factual_inconsistency_longdoc_2023,factscore_2023}. While effective in their respective settings, these approaches often provide limited entity-level evidence traceability in RAG pipelines or incur additional overhead.

\begin{figure*}[t]
  \centering
  \includegraphics[width=\textwidth]{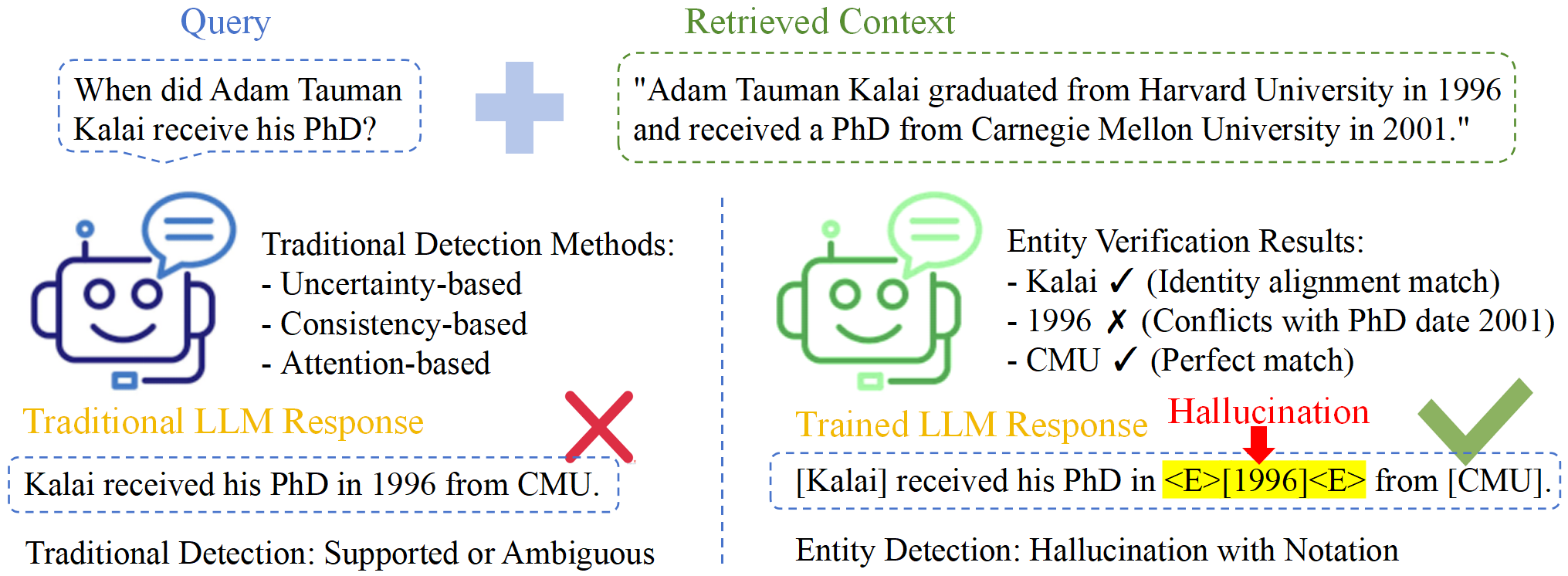}
  \caption{Comparison of traditional and RAG-based entity hallucination detection methods. Although the retrieved context contains correct factual evidence, traditional detection methods operating at the answer or sentence level may fail to identify subtle factual errors, whereas entity-level verification explicitly aligns generated entities with evidence and reveals fine-grained hallucinations. Example adapted from \citep{kalai2025language}.}
  \label{fig:motivation}
\end{figure*}


However, traditional detection approaches face fundamental limitations when deployed in real-world applications. As illustrated in Figure~\ref{fig:motivation}, models often generate hallucinations about recent events, specialized domains, or rapidly evolving information that falls outside their training data coverage~\citep{mallen2023not}. Additionally, reliance on internal model signals makes these methods vulnerable to distribution shifts and domain-specific biases that can compromise detection reliability. To mitigate coverage and recency limitations, RAG augments generation with retrieved evidence---making evidence explicitly available and shifting hallucination detection toward evidence-traceable verification~\citep{lewis2020rag,gao2023retrieval,Yeh2025HalluEntity}. RAG systems dynamically incorporate relevant documents during generation, enabling models to access up-to-date information while providing explicit evidence for factual claims.

Despite the promise of evidence-grounded generation, RAG introduces new requirements for hallucination detection. Models can still fabricate entities even when correct information exists within the retrieved context, leading to misalignments between evidence and generation~\citep{niu2024ragtruth}. Moreover, many recent RAG faithfulness detectors rely on external judges for verification, which incur additional complexity and potential error propagation, and often operate at token-, sentence-, or paragraph-level granularity. Such coarse-grained decisions fail to localize the entity-centric factual commitments that users most critically verify~\citep{yue2023automatic}. Beyond granularity, evidence-based detection in RAG must also be robust to \emph{spurious correlations}, where hallucinated entities appear supported due to superficial keyword overlaps in retrieved passages. Together, these issues highlight the need for entity-level, evidence-traceable verification that can distinguish genuine factual support from accidental matches. This motivates our central question:

\begin{center}
\textbf{\textit{How can we leverage RAG to enhance hallucination detection by establishing direct entity--evidence alignment in retrieved contexts?}}
\end{center}

Building on this foundation, we propose \textit{Evidence-Aligned Entity Verification} (EAEV), an entity-anchored and evidence-traceable hallucination detector that operates entirely within retrieved contexts. EAEV verifies each entity mention through complementary identity, semantic, and consistency alignment, and further incorporates counterfactual stability analysis to distinguish robust evidence support from fragile, surface-level matches. By grounding verification at the entity level, EAEV enables fine-grained localization of factual errors while remaining robust to spurious correlations. Extensive experiments demonstrate the effectiveness of EAEV, achieving 87.89\% AUROC on LLaMA2-13B with strong generalization across datasets. Our main contributions are summarized as follows:

\begin{itemize}
  \item We establish \textit{RAG-based hallucination detection} (RHD) as a practically important setting for hallucination detection in RAG pipelines, formulating it as \textit{entity-level evidence alignment} verification to provide fine-grained, evidence-traceable decisions beyond prior methods based on uncertainty estimation or black-box judgment signals.
  \item We propose \textit{Evidence-Aligned Entity Verification} (EAEV), a novel method that combines multi-dimensional alignment with counterfactual stability analysis to distinguish genuine evidence support from spurious correlations.
  \item We demonstrate superior performance and generalization across multiple RAG benchmarks and model architectures, achieving state-of-the-art results while maintaining practical deployability.
\end{itemize}

\section{Learning Setups}
\label{sec:Learning Setups}

In this section, we present necessary notations and establish the theoretical foundation for RAG-based hallucination detection, emphasizing entity-centric verification within retrieved contexts.

\subsection{Basic Definitions}

Following previous work \cite{oh2025understanding,du2024haloscope}, we use a distribution
${P}_{\boldsymbol{\theta}}(\cdot)$ over token sequences to define LLM,
where $\boldsymbol{\theta}$ is the model parameters.
Given a token sequence $\mathbf{Q} = [x_1, \dots, x_k]$ representing the query,
where each $x_i$ is the $i$-th token in the sequence.
${P}_{\boldsymbol{\theta}}(\cdot)$ generates an answer
$\mathbf{A} = [x_{k+1}, \dots, x_{k+q}]$
by predicting each token based on the preceding context: $P_{\boldsymbol{\theta}}(x_i \mid x_1, \dots, x_{i-1})$, for $i = k+1, \dots, k+q$.


\subsection{Traditional Hallucination Detection}

Traditional hallucination detection aims to identify incorrect content in LLM outputs. Given a query $\mathbf{Q}$ and an answer $\mathbf{A}$, a detector $D$ produces $\hat{y} = D(\mathbf{Q}, \mathbf{A})$, where $\hat{y} \in \{0, 1\}$ indicates hallucination. Existing methods operate through uncertainty estimation, consistency checking, or external verification, but face challenges when evidence is available yet underutilized in RAG settings.

\subsection{RAG-based Hallucination Detection}

RHD represents a fundamental shift from traditional approaches by leveraging retrieved evidence for verification. Unlike conventional methods that rely solely on model internals, RHD operates under the assumption that factual accuracy can be determined through explicit alignment between generated content and available evidence within retrieved contexts $\mathcal{P}$. We formalize RHD as follows:
given a query $\mathbf{Q}$, retrieved contexts $\mathcal{P}$, and a generated answer $\mathbf{A}$, the objective of RHD is to learn a detector $D$ that determines factual accuracy through evidence alignment:
\begin{equation}
D(\mathbf{Q}, \mathbf{A}, \mathcal{P}) =
\begin{cases}
1, & \text{if } \mathbf{A} \text{ is supported} \\
   & \text{by evidence in } \mathcal{P}, \\
0, & \text{otherwise}.
\end{cases}
\end{equation}

The key insight is that factual errors in RAG settings manifest primarily at the entity level, where specific named entities, temporal expressions, and quantities determine overall response reliability.

\subsection{Entity-Centric Verification Framework}

For entity-centric verification, we extract candidate mentions $s$ from the generated answer $\mathbf{A}$, where each mention has type
$t \in \{\text{ENT}, \text{NUM}, \text{NP}\}$ corresponding to named entities, numerical values, and noun phrases. For each mention $s$, we retrieve evidence windows from $\mathcal{P}$ and select primary evidence $e^*$ through relevance scoring. We define three core alignment functions: identity alignment $\text{Id}(s,e^*) \in [0,1]$ measuring surface correspondence, semantic alignment $\text{Sem}(s,e^*) \in [-1,1]$ capturing meaning preservation, and consistency alignment $\text{Con}(s,e^*) \in [0,1]$ evaluating quantitative agreement and conflict detection.

For each mention $s$, we compute support signals through weighted combination of alignment dimensions and detect conflicts through binary indicators. To distinguish robust evidence from spurious correlations, we apply counterfactual stability analysis using perturbation sets $\mathcal{U}$. Finally, mentions are aggregated into entity-level decisions through canonicalization, producing interpretable verification scores with direct evidence traceability.

Due to space constraints, the related work is discussed in Appendix~\ref{Related Works}.

\section{Methodology}
\label{sec:methodology}

\subsection{Motivation and Observations}

Effective RAG verification requires understanding how factual errors manifest in the presence of relevant evidence. As illustrated in Figure~\ref{fig:motivation}, traditional hallucination detection methods rely solely on internal model signals and are limited by training data coverage, while our RAG-enhanced approach incorporates external evidence sources to improve detection accuracy and coverage. Consider a model that is provided with partially correct supporting documents, as shown in the OpenAI example in Figure~\ref{fig:motivation}~\citep{kalai2025language}. This example illustrates a fundamental challenge: \textit{models can fabricate specific entities while correctly incorporating other factual elements from the context}, as noted by recent analysis of why language models hallucinate.

Existing detection methods operating at sentence or paragraph levels fail to localize such precise factual inconsistencies, as the overall semantic coherence remains high despite the critical entity-level error. Prior analyses suggest that hallucinations often manifest as entity-centric errors (e.g., names, dates, quantities), which are particularly important for users' trust judgments~\citep{Yeh2025HalluEntity}. This observation motivates our entity-centric approach: rather than evaluating global semantic consistency, we decompose verification into atomic factual units where evidence alignment can be precisely established and traced.

\subsection{Framework Overview}

\begin{figure*}[t]
  \centering
  \includegraphics[width=\textwidth]{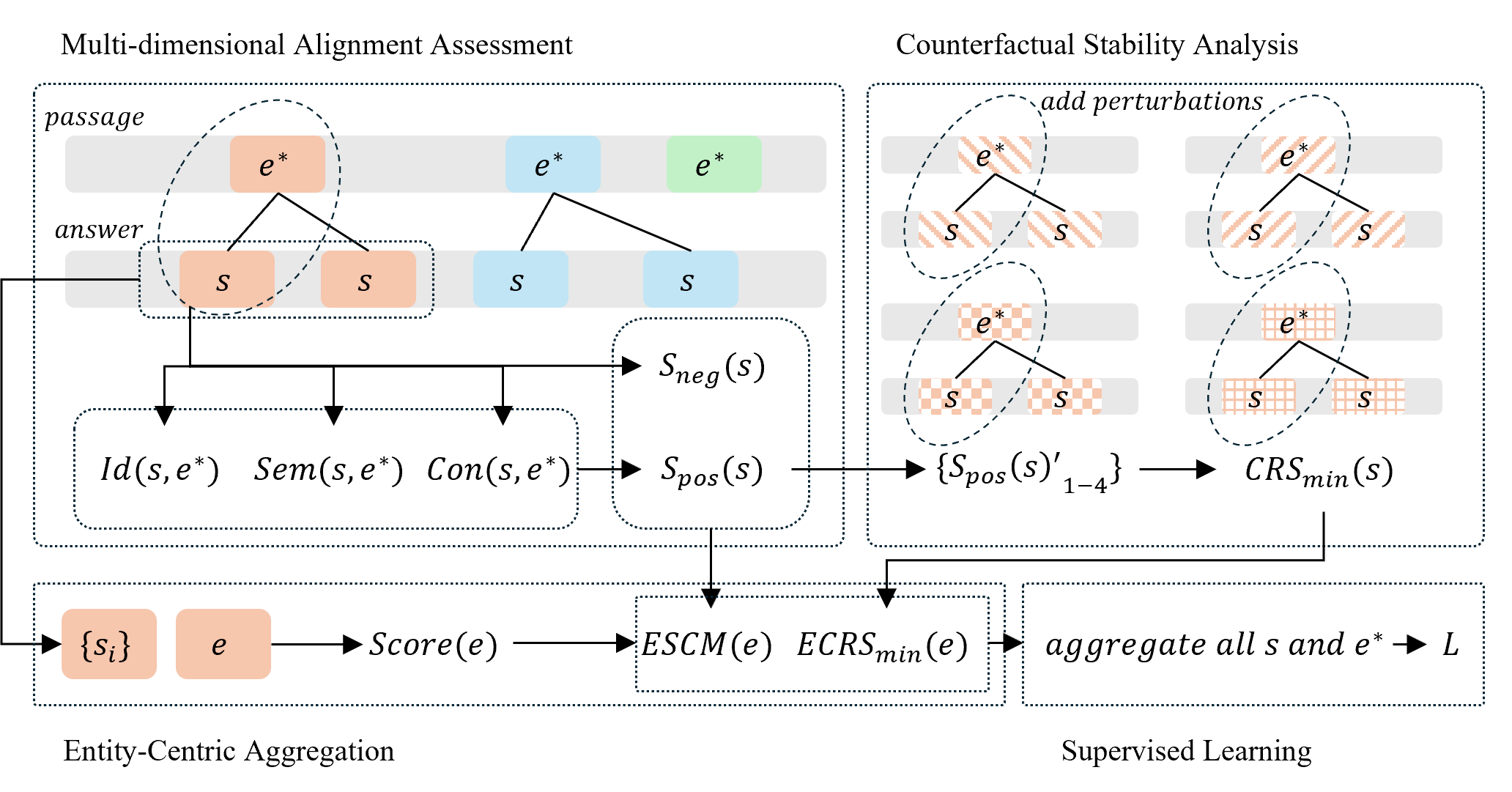}
  \caption{Overview of Evidence-Aligned Entity Verification (EAEV), which leverages RAG to align generated entities with retrieved evidence contexts, evaluates entity--evidence correspondence through multi-dimensional alignment and counterfactual stability analysis, and aggregates these signals to supervise a unified verifier.}
  \label{fig:methodology}
\end{figure*}

To address these challenges, we propose EAEV, which transforms entity verification into a systematic evidence alignment task through four interconnected stages that maintain evidence traceability throughout verification. As shown in Figure~\ref{fig:methodology}, the framework operates under three core principles: context-only verification where all signals derive from alignment between generated content and retrieved evidence, entity-centric aggregation enabling cross-mention evidence consolidation, and unified verification architecture that produces interpretable signals and supports supervised fine-tuning for deployment.

Given a query $\mathbf{Q}$, retrieved context $\mathcal{P}$, and generated answer $\mathbf{A}$, we first perform candidate mention extraction to identify factual commitments in $\mathbf{A}$ and construct local answer windows, followed by evidence retrieval and selection that identifies relevant evidence windows from $\mathcal{P}$ and selects primary evidence $e^*$ for each mention. These steps provide the necessary inputs for verification. EAEV then performs entity-level verification through four core stages: 
(1) \textit{multi-dimensional alignment assessment}, which evaluates entity--evidence correspondence through complementary identity, semantic, and consistency signals; 
(2) \textit{counterfactual stability analysis}, which tests the robustness of evidence alignment under controlled perturbations to distinguish genuine support from spurious correlations; 
(3) \textit{entity-centric aggregation}, which consolidates mention-level signals into entity-level decisions with explicit evidence traceability; and 
(4) \textit{EAEV-guided supervised learning}, which transfers these verification signals into a unified verifier through standard fine-tuning.
This modular design enables fine-grained, interpretable hallucination detection while maintaining practical deployability.

\subsection{Multi-Dimensional Alignment Assessment}

For each candidate mention $s$ extracted from the answer and its selected primary evidence $e^*$ from the retrieved context, we evaluate alignment through three complementary dimensions, as different types of factual support fail in distinct ways and cannot be reliably captured by a single alignment signal.

\subsubsection{Identity Alignment}

Identity alignment captures direct matches through lexical forms and aliases, providing precise signals for exact alignment. This dimension uses a normalized similarity function that blends exact substring matching with fuzzy token-level matching:
\begin{equation}
\begin{split}
&\text{Id}(s,e^*)\\ = &\max\left(\mathbb{I}[s \subseteq e^* \vee e^* \subseteq s], \text{TSR}(s,e^*)\right),
\end{split}
\end{equation}
where $\mathbb{I}[\cdot]$ is the indicator function and $\text{TSR}(s,e^*) \in [0,1]$ computes the normalized token set ratio measuring lexical overlap between mention and evidence tokens. This formulation prioritizes exact matches while gracefully handling orthographic variations and aliases through the fuzzy matching fallback, ensuring robust identity detection across diverse lexical forms.

\subsubsection{Semantic Alignment}

Semantic alignment evaluates meaning preservation through embedding similarity, capturing paraphrases and reformulations that maintain factual content despite variations:
\begin{equation}
\text{Sem}(s,e^*) = \cos(f_{\text{enc}}(s), f_{\text{enc}}(e^*)),
\end{equation}
where $f_{\text{enc}}(\cdot)$ represents sentence-level embedding encoding that captures semantic correspondence beyond explicit textual correspondence. This approach enables detection of semantically equivalent expressions while maintaining computational efficiency, though it requires careful calibration to prevent accepting spurious semantic matches that lack genuine factual grounding.

\subsubsection{Consistency Alignment}

Consistency alignment addresses value correspondence and explicit factual conflicts through numerical overlap assessment combined with lightweight pattern-based contradiction detection. For mentions with quantitative attributes, we measure value consistency via normalized intersection over union of extracted numerical values:
\begin{equation}
\text{Con}(s,e^*) = \frac{|N(s) \cap N(e^*)|}{|N(s) \cup N(e^*)|},
\end{equation}
where $N(\cdot)$ extracts and normalizes numerical values from text. We additionally define an \emph{anchor} indicator $A(\mathbf{Q},e^*)\in\{0,1\}$, which is $1$ if the selected evidence contains key terms from the original query. Moreover, we detect explicit contradictions through $S_{\text{neg}}(s) \in \{0,1\}$ using lightweight pattern matching that identifies temporal mismatches, numerical conflicts, and relational inconsistencies, providing high-precision negative signals that complement positive support.

\subsubsection{Type-Adaptive Support Synthesis}
We synthesize the three alignment dimensions into a unified support score that adapts to mention types, recognizing that different mention categories rely on different verification signals. For each mention $s$ of type $t \in \{\text{ENT}, \text{NUM}, \text{NP}\}$, we compute the support signal $S_{\text{pos}}(s) \in [0,1]$ as:
\begin{equation}
\begin{aligned}
&\;
w_I^{(t)} \cdot \text{Id}(s,e^*)
+ w_S^{(t)} \cdot \text{Sem}(s,e^*) \\
+ &w_C^{(t)} \Big(
\text{Con}(s,e^*) + b_{\text{anc}} \cdot A(\mathbf{Q},e^*)
\Big),
\end{aligned}
\end{equation}
where type-adaptive weights $w_I^{(t)}, w_S^{(t)}, w_C^{(t)}$ are optimized for each mention type---numerical mentions emphasize consistency alignment while named entities prioritize identity and semantic alignment. We incorporate the contradiction indicator $S_{\text{neg}}(s)\in\{0,1\}$ to compute the support margin $\text{SCM}(s)=S_{\text{pos}}(s)-\beta\cdot S_{\text{neg}}(s)$, where $\beta$ controls conflict penalties.

\subsection{Counterfactual Stability Analysis}


A critical challenge in RAG-based verification is that surface-level evidence alignment alone is insufficient to establish factual support, as hallucinated content can accidentally match retrieved text through spurious correlations. Traditional alignment metrics can be deceived by coincidental keyword overlaps or formatting artifacts that create false signals of factual support. To address this limitation, we introduce counterfactual stability analysis, which treats robustness under controlled perturbations as a necessary condition for evidence-based verification.

The core insight is that genuine factual correspondence should persist across minor variations in text presentation, while spurious matches are inherently fragile and collapse when surface features change. We define perturbation sets $\mathcal{U}$ containing controlled variations that preserve semantic content while altering surface characteristics. For each mention $s$, we compute stability bounds: minimum support $\text{CRS}_{\min}(s) = \min_{u \in \mathcal{U}} S_{\text{pos}}^{(u)}(s)$ measuring the lowest support under perturbations, and stability gaps $\text{CRS}_{\Delta}(s) = S_{\text{pos}}(s) - \text{CRS}_{\min}(s)$ indicating robustness to variations.

We instantiate $\mathcal{U}$ with four targeted perturbations that address distinct sources of spurious correlation: (1) \textit{leave-one-out evidence removal} eliminates the strongest evidence window to test dependency on single sources, preventing over-reliance on potentially misleading context; (2) \textit{punctuation and case normalization} removes formatting artifacts and capitalization patterns creating false lexical matches, ensuring alignment reflects genuine content rather than presentation; (3) \textit{whitespace compression} eliminates spacing variations and tokenization inconsistencies that might artificially inflate similarity scores; and (4) \textit{alphanumeric-only filtering} retains only core semantic content by removing symbols and special characters that could create spurious token-level alignments.

High minimum support $\text{CRS}_{\min}(s)$ indicates that evidence alignment persists across these controlled variations, indicating the factual correspondence is robust rather than circumstantial. Conversely, large stability gaps $\text{CRS}_{\Delta}(s)$ reveal fragile correlations that depend on specific textual configurations, flagging potentially unreliable evidence support. This stability analysis enables EAEV to distinguish authentic factual grounding from accidental surface-level matches, improving detection precision in challenging cases where traditional alignment metrics alone prove insufficient.

\subsection{Entity-Centric Aggregation}

For entities with multiple mentions across the answer, we consolidate evidence signals to obtain robust entity-level assessments. We canonicalize mention strings through lowercasing, punctuation and article removal to identify coreferent mentions, and apply lightweight pronoun resolution that links pronouns to the most recent non-pronoun entity. 

For an entity $e$ with mention set $\{s_i\}$, we aggregate verification signals conservatively. Positive support uses top-K averaging $\text{ES}_{\text{pos}}(e) = \text{mean}(\text{topK}\{S_{\text{pos}}(s_i)\})$ to emphasize strongest evidence across mentions. Negative signals use max pooling $\text{ES}_{\text{neg}}(e) = \max\{S_{\text{neg}}(s_i)\}$ for conservative conflict detection, ensuring any mention-level conflict propagates to entity level. Stability becomes $\text{ECRS}_{\min}(e) = \min\{\text{CRS}_{\min}(s_i)\}$ to identify the weakest link across all entity mentions.

The entity consistency margin $\text{ESCM}(e) = \text{ES}_{\text{pos}}(e) - \beta_e \cdot \text{ES}_{\text{neg}}(e)$ integrates these consolidated signals, where $\beta_e$ controls entity-level conflict penalties. The final entity verification score integrates consistency and stability through a multiplicative combination:
\begin{equation*}
\text{score}(e) = \sigma(-\text{ESCM}(e)) \cdot (1 - \sigma(\text{ECRS}_{\min}(e)))
\end{equation*}
where $\sigma(\cdot)$ denotes the sigmoid function. This formulation produces high risk scores for entities with weak evidence support or low stability, enabling answer-level assessment through max pooling over entity scores while preserving traceability to specific evidence windows. In our main setting, these alignment and stability scores are used to construct supervision for training a single-model verifier, all evaluations use the fine-tuned model for inference.

\subsection{EAEV-Guided Supervised Learning}

The alignment and stability signals computed by EAEV provide direct supervision for training models to perform interpretable hallucination detection through entity-level annotation. Rather than requiring complex architectural modifications, we leverage EAEV's verification capabilities to construct high-quality training data where models learn to reproduce answers while marking unsupported entities with verification tags.

For each training instance, we generate target sequences where entities with $\text{ESCM}(e) < \tau_{\text{threshold}}$ are enclosed in $\langle\text{E}\rangle$ markers, creating supervision that directly transfers EAEV's multi-dimensional verification logic to generation. We optimize a token-weighted cross-entropy that transfers EAEV's entity-level signals into generation:
\begin{equation}
L = \sum_{t} w_t \cdot \text{CE}(p_{\boldsymbol{\theta}}(y_t|x,y_{<t}), y_t),
\end{equation}
where
\begin{equation}
\begin{aligned}
r_t &= \max_{t \in e}\sigma\big(-\text{ESCM}(e)\big), \\
u_t &= \max_{t \in e}\big(1-\sigma(\text{ECRS}_{\min}(e))\big),\\
w_t &= \text{clip}\big(1 + \alpha r_t + \gamma u_t,\; w_{\min}, w_{\max}\big),
\end{aligned}
\end{equation}
and tokens outside any tagged entity use $\max_{t \in e} = 0$.

This approach enables standard supervised fine-tuning to learn EAEV's verification patterns, transferring interpretable entity-level detection capabilities into generation without requiring specialized decoding procedures or multi-model coordination. In all main experiments, the reported results of \emph{EAEV} are obtained by fine-tuning the backbone LLM with the EAEV-guided supervised learning procedure described in this subsection.

\section{Experiment}

\subsection{Experiment Settings}

We evaluate EAEV on three representative benchmarks: RAGTruth~\citep{niu2024ragtruth}, HotpotQA~\citep{yang2018hotpotqa}, and DelucionQA~\citep{sadat-etal-2023-delucionqa}. Experiments are conducted with three LLM backbones: Qwen2.5-7B~\citep{yang2024qwen25}, LLaMA2-7B~\citep{touvron2023llama2}, and LLaMA2-13B~\citep{touvron2023llama2} and compared against multiple strong hallucination detection baselines, including SelfCheckGPT~\citep{manakul2023selfcheckgpt}, Semantic Entropy~\citep{kuhn2023semanticuncertainty}, LLM-Check~\citep{jain2024llmcheck}, EarlyDetect~\citep{snyder2024earlydetect}, NoVo~\citep{ho2025novo}, RAGAS~\citep{es2024ragas}, RefChecker~\citep{hu2024refchecker}, ReDEeP~\citep{sun2024redeep}, TSV~\citep{park2025tsv}, Linear Probe~\citep{duan2024hiddenstates}, and HaloScope~\citep{du2024haloscope}. 
We employ three complementary metrics for evaluation: area under the receiver operating characteristic curve (AUROC), Accuracy, and F1 score. Full experimental details are provided in Appendix~\ref{Experimental_Details}.

\subsection{Experimental Results and Analysis}

\subsubsection{Main Results}
EAEV achieves strong performance across all evaluation settings, demonstrating the effectiveness of entity-centric evidence alignment for RAG hallucination detection. As shown in Table~\ref{tab:main_results}, our method attains 84.72\% average AUROC on Qwen2.5-7B, 79.63\% on LLaMA2-7B, and 87.55\% on LLaMA2-13B, representing improvements of 2.29, 2.59, and 3.34 percentage points respectively over the strongest baseline TSV. These gains across different backbone architectures support our core hypothesis that entity-level factual errors constitute a model-agnostic challenge in RAG systems. Larger models exhibit higher performance, with LLaMA2-13B achieving the best results, suggesting that EAEV can effectively leverage increased model capacity for more accurate evidence alignment.

Cross-dataset evaluation shows that EAEV generalizes across diverse reasoning scenarios and domain requirements. On RAGTruth, the method demonstrates strong sensitivity to nuanced factual inconsistencies in general knowledge contexts, while results on HotpotQA and DelucionQA indicate effective verification in multi-hop reasoning and precision-critical technical domains. Together, these results suggest that the proposed multi-dimensional alignment framework captures broadly applicable patterns in entity-level hallucination detection across tasks with distinct characteristics.


The observed scaling behavior highlights the practical applicability of EAEV across different deployment regimes. While larger models benefit from increased capacity for sophisticated evidence alignment, smaller models also achieve meaningful improvements, enabling use in resource-constrained settings. Overall, these findings indicate that EAEV provides a scalable and practical direction for hallucination detection under varied model sizes and task demands.

\begin{table*}[tb]
  \centering
  \small
  \setlength{\tabcolsep}{4pt}

  \begin{tabular}{l ccc ccc ccc ccc}
    \toprule
    & \multicolumn{3}{c}{\textbf{RAGTruth}}
    & \multicolumn{3}{c}{\textbf{HotpotQA}}
    & \multicolumn{3}{c}{\textbf{DelucionQA}}
    & \multicolumn{3}{c}{\textbf{Average}} \\
    \cmidrule(lr){2-4}\cmidrule(lr){5-7}\cmidrule(lr){8-10}\cmidrule(lr){11-13}
    \textbf{Method}
    & AUROC & Acc & F1
    & AUROC & Acc & F1
    & AUROC & Acc & F1
    & AUROC & Acc & F1 \\
    \midrule

    \multicolumn{13}{c}{\textit{Qwen2.5-7B}} \\
    \midrule
    EarlyDetect      & 66.38 & 70.12 & 65.34 & 67.15 & 69.32 & 65.10 & 68.25 & 69.83 & 66.21 & 67.26 & 69.76 & 65.55 \\
    Selfcheckgpt     & 64.39 & 69.05 & 63.28 & 65.73 & 68.32 & 63.95 & 66.43 & 68.76 & 65.01 & 65.52 & 68.71 & 64.08 \\
    Novo             & 73.79 & 76.21 & 68.67 & 74.85 & 75.75 & 69.12 & 76.03 & 76.12 & 70.43 & 74.89 & 76.03 & 69.41 \\
    Linear Probe     & 75.27 & 77.38 & 69.52 & 76.11 & 77.02 & 69.84 & 77.10 & 77.54 & 71.02 & 76.16 & 77.31 & 70.13 \\
    HaloScope        & 71.01 & 74.16 & 67.70 & 72.24 & 73.90 & 68.20 & 73.01 & 74.16 & 69.05 & 72.09 & 74.07 & 68.32 \\
    LLM-Check        & 62.75 & 61.07 & 62.18 & 63.91 & 67.20 & 62.93 & 65.14 & 67.86 & 64.23 & 63.93 & 65.38 & 63.11 \\
    Semantic Entropy & 65.43 & 70.65 & 64.71 & 66.87 & 69.83 & 65.25 & 68.02 & 70.19 & 65.84 & 66.77 & 70.22 & 65.27 \\
    RAGAS            & 74.76 & 77.32 & 69.90 & 76.02 & 76.84 & 70.40 & 76.89 & 77.01 & 71.43 & 75.89 & 77.06 & 70.58 \\
    RefCheck         & 73.25 & 75.89 & 68.43 & 74.61 & 75.30 & 68.81 & 75.20 & 75.76 & 70.01 & 74.35 & 75.65 & 69.08 \\
    ReDEeP           & 77.87 & 78.51 & 71.92 & 79.43 & 78.23 & 72.74 & 80.12 & 78.96 & 73.52 & 79.14 & 78.57 & 72.73 \\
    TSV              & 79.34 & 77.69 & 72.37 & 82.07 & 79.36 & 72.28 & \textbf{85.87} & \textbf{80.47} & 74.97 & 82.43 & 79.17 & 73.21 \\
    \rowcolor{myteal!30}
    \textbf{EAEV (Ours)}
                     & \textbf{81.73} & \textbf{80.04} & \textbf{74.28}
                     & \textbf{86.74} & \textbf{81.23} & \textbf{74.35}
                     & 85.68 & 80.25 & \textbf{75.62}
                     & \textbf{84.72} & \textbf{80.51} & \textbf{74.75} \\
    \midrule

    \multicolumn{13}{c}{\textit{LLaMA2-7B}} \\
    \midrule
    EarlyDetect      & 65.12 & 68.87 & 63.98 & 66.23 & 68.05 & 63.55 & 67.12 & 68.42 & 64.33 & 66.16 & 68.45 & 63.95 \\
    Selfcheckgpt     & 63.18 & 67.31 & 62.01 & 64.45 & 66.25 & 62.74 & 65.37 & 66.90 & 63.45 & 64.33 & 66.82 & 62.73 \\
    Novo             & 72.25 & 70.79 & 67.12 & 73.31 & 74.82 & 67.62 & 74.38 & 72.13 & 68.92 & 73.31 & 72.58 & 67.89 \\
    Linear Probe     & 73.56 & 76.36 & 68.23 & 74.25 & 75.43 & 68.55 & 75.48 & 75.62 & 70.02 & 74.43 & 75.80 & 68.93 \\
    HaloScope        & 69.83 & 73.24 & 66.31 & 70.92 & 72.43 & 66.75 & 71.74 & 73.25 & 67.45 & 70.83 & 72.97 & 66.84 \\
    LLM-Check        & 61.47 & 60.42 & 60.08 & 62.63 & 65.32 & 61.32 & 63.71 & 66.19 & 62.03 & 62.60 & 63.98 & 61.14 \\
    Semantic Entropy & 64.12 & 69.01 & 63.43 & 65.25 & 68.32 & 63.78 & 66.30 & 69.02 & 64.01 & 65.22 & 68.78 & 63.74 \\
    RAGAS            & 73.11 & 76.25 & 68.11 & 74.43 & 75.62 & 68.75 & 75.45 & 76.12 & 70.12 & 74.33 & 76.00 & 68.99 \\
    RefCheck         & 71.66 & 74.83 & 66.92 & 73.08 & 74.41 & 67.30 & 74.02 & 74.88 & 68.15 & 72.92 & 74.71 & 67.46 \\
    ReDEeP           & 76.42 & 78.01 & 71.23 & 74.58 & 77.15 & 71.94 & 78.35 & 77.66 & 72.43 & 76.45 & 77.61 & 71.87 \\
    TSV              & 78.87 & 77.02 & 72.01 & 75.04 & 78.83 & 72.01 & 77.21 & 77.52 & 73.95 & 77.04 & 77.79 & 72.66 \\
    \rowcolor{myteal!30}
    \textbf{EAEV (Ours)}
                     & \textbf{82.12} & \textbf{80.12} & \textbf{74.07}
                     & \textbf{77.91} & \textbf{80.56} & \textbf{73.39}
                     & \textbf{78.87} & \textbf{78.32} & \textbf{74.45}
                     & \textbf{79.63} & \textbf{79.67} & \textbf{73.97} \\
    \midrule

    \multicolumn{13}{c}{\textit{LLaMA2-13B}} \\
    \midrule
    EarlyDetect      & 67.18 & 70.01 & 65.12 & 68.42 & 69.05 & 65.87 & 69.66 & 69.81 & 66.50 & 68.42 & 69.62 & 65.83 \\
    Selfcheckgpt     & 65.47 & 68.10 & 63.02 & 66.93 & 67.22 & 63.89 & 67.82 & 67.88 & 64.52 & 66.74 & 67.73 & 63.81 \\
    Novo             & 80.54 & 76.35 & 69.31 & 78.68 & 76.13 & 70.16 & 77.12 & 76.55 & 71.22 & 78.78 & 76.34 & 70.23 \\
    Linear Probe     & 76.31 & 77.65 & 70.54 & 77.54 & 77.33 & 71.39 & 78.82 & 77.97 & 72.82 & 77.56 & 77.65 & 71.58 \\
    HaloScope        & 71.74 & 74.45 & 68.20 & 72.81 & 73.25 & 68.91 & 73.93 & 74.15 & 69.85 & 72.83 & 73.95 & 68.99 \\
    LLM-Check        & 63.83 & 67.12 & 61.95 & 65.28 & 64.62 & 62.73 & 66.55 & 66.98 & 63.45 & 65.22 & 66.24 & 62.71 \\
    Semantic Entropy & 66.02 & 70.20 & 64.62 & 67.35 & 69.02 & 65.21 & 68.40 & 70.10 & 65.83 & 67.26 & 69.77 & 65.22 \\
    RAGAS            & 75.67 & 73.63 & 70.22 & 76.88 & 77.01 & 71.01 & 78.02 & 77.66 & 72.52 & 76.86 & 76.10 & 71.25 \\
    RefCheck         & 74.25 & 76.00 & 68.85 & 75.66 & 75.43 & 69.33 & 76.92 & 76.41 & 70.44 & 75.61 & 75.95 & 69.54 \\
    ReDEeP           & 82.44 & 78.39 & 74.35 & 81.39 & 78.43 & 73.21 & 81.04 & 79.81 & 74.15 & 81.62 & 78.88 & 73.90 \\
    TSV              & 84.55 & 80.12 & 73.50 & 83.12 & 79.43 & 73.22 & 84.96 & 80.12 & 75.68 & 84.21 & 79.89 & 74.13 \\
    \rowcolor{myteal!30}
    \textbf{EAEV (Ours)}
                     & \textbf{87.89} & \textbf{84.29} & \textbf{76.85}
                     & \textbf{88.12} & \textbf{83.53} & \textbf{75.59}
                     & \textbf{86.65} & \textbf{83.22} & \textbf{77.92}
                     & \textbf{87.55} & \textbf{83.68} & \textbf{76.79} \\
    \bottomrule
  \end{tabular}
  \caption{Performance comparison across different models and datasets. All results are averaged over three independent runs, with the rightmost columns showing metrics averaged across datasets.}
  \label{tab:main_results}
\end{table*}

\subsubsection{Ablation Study}

\begin{figure}[t]
  \centering
  \includegraphics[width=\columnwidth]{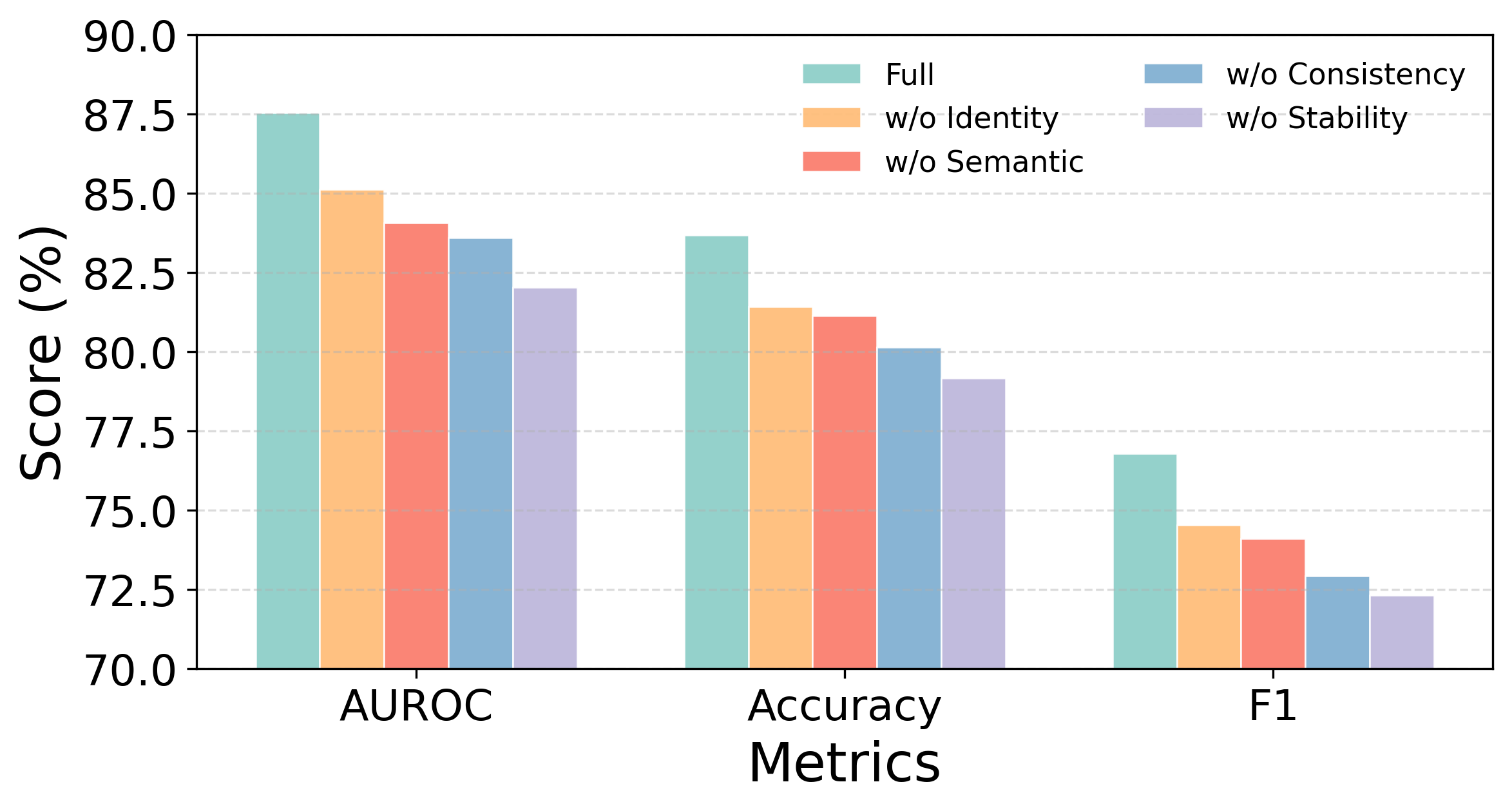}
  \caption{Ablation study results on LLaMA2-13B, showing ablation analysis on different components.}
  \label{fig:ablation}
\end{figure}

To validate each component's contribution, we conduct comprehensive ablation studies across all benchmarks. As shown in Figure~\ref{fig:ablation}, counterfactual stability analysis provides the most substantial contribution, confirming the necessity of our approach for distinguishing genuine evidence support from spurious correlations. The results demonstrate that each alignment dimension contributes meaningfully to overall performance, with balanced degradation patterns indicating that all components address distinct verification challenges. The full framework consistently achieves the best performance, indicating that the proposed components jointly contribute to effective verification. More ablation details are provided in Appendix~\ref{sec:ablation_details}.

\subsubsection{Sensitivity Analysis}

\begin{figure}[t]
  \centering
  \includegraphics[width=\columnwidth]{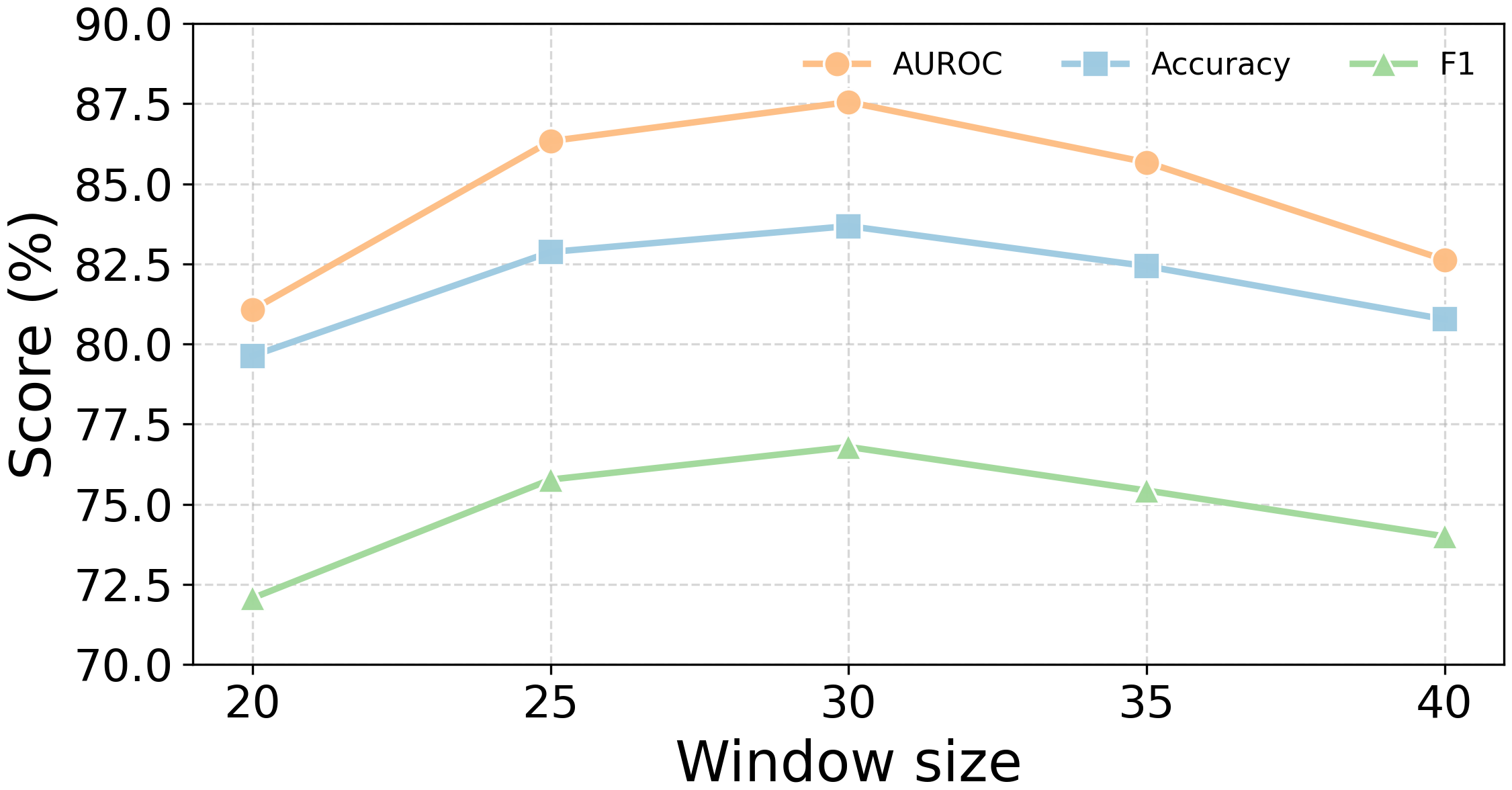}
  \caption{Ablation study results on LLaMA2-13B, showing ablation analysis on window size sensitivity.}
  \label{fig:window}
\end{figure}

We analyze EAEV's robustness to answer-side window length, a key parameter controlling contextual span during evidence alignment. As shown in Figure~\ref{fig:window}, the framework achieves optimal performance with 30-token windows while maintaining stability across the practical range. Smaller windows limit contextual information for accurate alignment, while larger windows introduce noise that dilutes alignment signals. The framework demonstrates reasonable robustness within the 25-35 token range, validating our parameter choice and confirming consistent performance across deployment scenarios. This analysis establishes EAEV's reliability and practical applicability under varying configuration settings. Details are provided in Appendix~\ref{sec:sensitivity_details}.



\section{Conclusion}


Hallucination detection remains critical for reliable RAG system deployment in factual applications. We introduce EAEV, a framework that performs entity-level verification through multi-dimensional evidence alignment and counterfactual stability analysis. By distinguishing genuine factual support from spurious correlations, EAEV addresses key challenges in RAG-based verification where hallucinated content may accidentally align with retrieved text. Experimental results demonstrate substantial improvements across benchmarks and model architectures. Importantly, EAEV operates entirely within retrieved contexts and produces interpretable verification signals, making it well aligned with practical RAG pipelines. Overall, these findings suggest that robust entity-level verification provides a practical and effective direction for improving hallucination detection in real-world evidence-grounded generation systems.

\section*{Limitations}
While our approach demonstrates strong effectiveness for RAG-based hallucination detection, several limitations should be acknowledged. First, LLM outputs can exhibit instability across different runs, particularly for complex queries that require multi-step reasoning. Second, as a context-only verification framework, EAEV depends on the quality and coverage of retrieved evidence; missing or incomplete retrieval may limit detection performance. Finally, while this work focuses on accurate and interpretable hallucination detection, it does not directly address mitigating hallucinations during generation. Integrating entity-level verification signals into decoding or training-time mitigation mechanisms is a promising direction for future work.

\section*{Acknowledgments}
Runsong Jia, Mengjia Wu, and Yi Zhang were supported by the Australian Commonwealth Scientific and Industrial Research Organization (CSIRO) in conjunction with the National Science Foundation (NSF) of the United States, under CSIRO-NSF \#2303037.
Zhen Fang was supported by grant DE250100363.


\bibliography{custom}

\appendix

\section{Appendix}
\label{sec:appendix}

\subsection{Related Works}
\label{Related Works}
\subsubsection{Traditional Hallucination Detection}
Traditional hallucination detection methods primarily leverage uncertainty estimation and self-consistency mechanisms within model outputs. Representative approaches include SelfCheckGPT, which measures semantic consistency across multiple generations~\citep{manakul2023selfcheckgpt}, and Semantic Entropy, which operates on meaning-level divergences~\citep{farquhar2024semantic}. Recent advances explore attention-level interpretability through NoVo~\citep{ho2025novo} and representational analysis of knowledge-awareness directions~\citep{ferrando2025doiknow}. While effective in controlled settings, these methods remain constrained by their reliance on internal model signals rather than explicit evidence verification.

\subsubsection{Evidence-Based Hallucination Detection in RAG}
RAG environments present unique challenges where hallucinations persist despite available evidence, motivating specialized detection approaches. RARR employs research and revision stages for evidence attribution and consistency-based correction~\citep{gao2023rarr}. FActScore provides atomic-level factual evaluation by decomposing generated text into verifiable claims~\citep{factscore_2023}. CoVe introduces systematic self-verification through question generation and independent answering~\citep{dhuliawala2024cove}. ReDeEP leverages mechanistic interpretability to disentangle parametric and contextual knowledge contributions~\citep{sun2024redeep}, while RAGTruth establishes evaluation infrastructure with fine-grained annotations~\citep{niu2024ragtruth}. These approaches highlight the importance of evidence-grounded verification but typically operate at coarse granularities or require external verification mechanisms. Our work addresses this limitation through entity-level verification within retrieved contexts, providing direct evidence traceability without dependencies on external judges.

\subsection{Experimental Details}
\label{Experimental_Details}
\subsubsection{Datasets Details}
\label{Datasets_Details}
\paragraph{RAGTruth}
RAGTruth provides a controlled environment for analyzing hallucinations in standard RAG pipelines. The corpus aggregates responses from both open-source and closed-source LLMs, accompanied by meticulous word-level manual annotations and instance-level labels across three task categories: question answering, data-to-text generation, and news summarization. The dataset comprises approximately 18,000 annotated responses in total. We compute all evaluation metrics at the answer level to maintain consistency across comparisons~\citep{niu2024ragtruth}.

\paragraph{RAGBench (HotpotQA \& DelucionQA)}
RAGBench is a large-scale benchmark containing approximately 100,000 examples with a standardized RAG schema that provides retrieved contexts and answer annotations suitable for hallucination detection tasks. The benchmark spans five domains and twelve component datasets. We utilize two representative components: (i) \textbf{HotpotQA}, a multi-hop question answering benchmark built on Wikipedia articles with sentence-level supporting facts that emphasizes cross-document reasoning capabilities; and (ii) \textbf{DelucionQA}, a domain-specific QA dataset constructed from automotive user manuals, featuring human-annotated labels that indicate whether answers contain hallucinations given the retrieved context. We adopt the benchmark's evaluation protocol and consistently assess performance at the answer level~\citep{friel2024ragbench,yang2018hotpotqa,sadat-etal-2023-delucionqa}.

\subsubsection{Implementation Details}
\label{sec:implementation_details}

We run all experiments on servers equipped with 4×NVIDIA A100 GPUs and server-grade multi-core processors. Our implementation is based on PyTorch and Hugging Face Transformers. We use LLaMA-Factory for LLM fine-tuning and inference (with LoRA). Unless otherwise specified, we employ greedy search for generation decoding, and all other parameters follow the default settings of each model.

For candidate construction and evidence retrieval, we retain at most 5 candidate windows per mention with $\text{top\_bm25}=2$ and $\text{top\_embed}=2$. Answer-side windows use $\text{window\_tokens}=30$ and $\text{stride}=15$. For multi-dimensional alignment, we use type-adaptive weights with defaults ENT: $(0.45, 0.45, 0.10)$, NUM: $(0.25, 0.25, 0.50)$, and NP: $(0.35, 0.35, 0.30)$. We set default $\beta = 1.0$ for consistency margin. For CRS analysis, we apply four perturbation types: leave-one-out, depunctuating and lowercasing, compressing whitespace, and retaining only alphanumeric characters. For entity grouping, we use conservative aggregation with default $K=2$ and $\beta_e = 1.0$. For decision rules, we scan $\tau_{\text{scm}} \in [-0.5, 0.2]$, $\tau_{\text{escm}} \in [-0.5, -0.1]$, $\tau_{\text{crs\_min}} \in [0.0, 0.5]$, and $K \in \{1, 2, 3\}$ on validation sets.

For EAEV-guided SFT, we insert $\langle E \rangle...\langle /E \rangle$ markers and select $(\alpha, \gamma, w_{\min}, w_{\max})$ on validation sets. Fine-tuning uses LLaMA-Factory with LoRA, following framework defaults except for token weighting and data annotation. All hyperparameters use validation set selection, and final results report best validation configurations.

\subsubsection{Evaluation Metrics}
Following prior works~\citep{kuhn2023semanticuncertainty,du2024haloscope}, we employ three complementary metrics to evaluate hallucination detection performance: area under the receiver operating characteristic curve (AUROC), Accuracy, and F1 score. 

\textbf{AUROC} measures the ability of a method to discriminate between truthful and hallucinated outputs across different decision thresholds. A higher AUROC indicates better overall ranking performance independent of a specific threshold. 

\textbf{Accuracy} is calculated by comparing predicted labels with ground-truth annotations under a fixed threshold (e.g., 0.5 on the similarity score between the generation and the reference). It reflects the proportion of correctly classified instances but can be biased when classes are imbalanced. 

\textbf{F1 score}, the harmonic mean of Precision and Recall, provides a balanced evaluation when both false positives and false negatives are costly. It is particularly useful in assessing detection performance under skewed class distributions. 

Together, these metrics ensure a comprehensive assessment of both ranking quality and classification reliability in hallucination detection.

\subsubsection{Model Details}
We conduct our experiments on three widely used large language models that represent different scales and training paradigms. \textbf{Qwen2.5-7B}~\citep{yang2024qwen25} is an open-source model from Alibaba’s Qwen series, designed with improved pre-training data and instruction tuning for multilingual reasoning. \textbf{LLaMA2-7B}~\citep{touvron2023llama2} and \textbf{LLaMA2-13B}~\citep{touvron2023llama2} are part of Meta’s LLaMA2 family, which have been extensively used as backbone models in academic research and industrial applications. Together, these models cover diverse capacities and training corpora, providing a representative testbed for evaluating hallucination detection methods.

\subsubsection{Details about Baseline Models}
\label{sec:baseline_details}
We compare our approach against eleven representative hallucination detection baselines. Below we briefly introduce each method and its underlying intuition.

\begin{itemize}
    \item \textbf{SelfCheckGPT}~\citep{manakul2023selfcheckgpt}: A zero-resource, sampling-based detector that repeatedly queries the model to generate multiple candidate responses and then measures their consistency. Greater inconsistency across samples suggests a higher risk of hallucination, making this method effective even without external evidence.
    
    \item \textbf{Semantic Entropy}~\citep{kuhn2023semanticuncertainty}: Estimates hallucination likelihood by computing linguistic invariances in token-level predictive distributions. When semantic alternatives diverge strongly in probability space, the model exhibits higher semantic entropy, indicating uncertainty and potential unreliability in factual grounding.
    
    \item \textbf{LLM-Check}~\citep{jain2024llmcheck}: Probes internal hidden states of LLMs with lightweight classifiers to directly flag hallucinations. By exploiting activation-level features, LLM-Check can detect subtle factual errors that do not manifest at the surface level but are encoded within the model’s latent representations.

    \item \textbf{Linear Probe}~\citep{duan2024hiddenstates}: A straightforward but effective baseline that trains linear classifiers on the hidden states of LLMs. By mapping internal activations to truthfulness labels, Linear Probe directly tests how much factuality information is encoded within raw model representations.
    
    \item \textbf{HaloScope}~\citep{du2024haloscope}: Leverages large quantities of unlabeled LLM outputs and applies energy-based and representation-driven detectors. By clustering semantic patterns across generations, HaloScope effectively identifies outliers that correspond to hallucinated claims with minimal supervision.
    
    \item \textbf{EarlyDetect}~\citep{snyder2024earlydetect}: A proactive detector that monitors generation in-progress. By analyzing partial outputs and their factual signals, EarlyDetect aims to catch hallucinations early, before the model produces fully misleading answers, thus enabling faster correction or intervention.
    
    \item \textbf{NoVo}~\citep{ho2025novo}: Stands for Norm Voting off hallucinations. This method measures the norms of attention heads and aggregates their “votes’’ to infer factual reliability. It leverages attention-level interpretability to highlight internal disagreement patterns that often precede hallucinated generations.
    
    \item \textbf{RAGAS}~\citep{es2024ragas}: Focuses on retrieval-augmented settings by breaking down model outputs into atomic statements and verifying each against retrieved passages. Faithfulness is quantified as the ratio of supported claims, allowing fine-grained detection of unsupported or fabricated content.
    
    \item \textbf{RefChecker}~\citep{hu2024refchecker}: Constructs structured knowledge graphs from model outputs and checks their alignment with external references. This graph-based perspective enables detection of hallucinations that may not be obvious at sentence level but become evident when relational consistency is examined.
    
    \item \textbf{ReDEeP}~\citep{sun2024redeep}: Employs mechanistic interpretability in retrieval-augmented generation (RAG). By tracing attention flow from queries to evidence passages, ReDEeP identifies whether the model’s factual claims are truly supported by retrieved documents or merely spurious correlations.
    
    \item \textbf{TSV}~\citep{park2025tsv}: Introduces the Truthfulness Separator Vector, which perturbs latent representations during inference to evaluate the stability of factual claims. Robust claims remain separable under perturbations, while hallucinated ones collapse, offering a novel perspective on truthfulness detection.
\end{itemize}

\subsection{Additional Results}

\subsubsection{Complete Ablation Study}
\label{sec:ablation_details}

\begin{table*}[t]

\begin{center}
\resizebox{\textwidth}{!}{
\begin{tabular}{l ccc ccc ccc ccc}
\toprule
\textbf{Variant}
& \multicolumn{3}{c}{\textbf{RAGTruth}}
& \multicolumn{3}{c}{\textbf{HotpotQA}}
& \multicolumn{3}{c}{\textbf{DelucionQA}}
& \multicolumn{3}{c}{\textbf{Average}} \\
\cmidrule(lr){2-4}\cmidrule(lr){5-7}\cmidrule(lr){8-10}\cmidrule(lr){11-13}
& AUROC & Acc & F1
& AUROC & Acc & F1
& AUROC & Acc & F1
& AUROC & Acc & F1 \\
\midrule
w/o Identity
& 84.97 & 82.72 & 74.23
& 86.12 & 81.35 & 73.20
& 84.26 & 81.82 & 75.89
& 85.12 & 81.43 & 74.53 \\
w/o Semantic
& 83.81 & 81.43 & 73.67
& 85.40 & 81.14 & 73.15
& 83.08 & 80.80 & 75.64
& 84.07 & 81.13 & 74.10 \\
w/o Consistency
& 85.67 & 82.13 & 74.55
& 83.42 & 79.37 & 71.08
& 81.85 & 78.96 & 73.34
& 83.60 & 80.15 & 72.93 \\
w/o Stability
& 82.49 & 79.84 & 72.47
& 82.30 & 79.06 & 71.65
& 81.73 & 78.70 & 73.06
& 82.03 & 79.17 & 72.33 \\
\midrule
\rowcolor{myteal!30}
\textbf{Full}
& \textbf{87.89} & \textbf{84.29} & \textbf{76.85}
& \textbf{88.12} & \textbf{83.53} & \textbf{75.59}
& \textbf{86.65} & \textbf{83.22} & \textbf{77.92}
& \textbf{87.55} & \textbf{83.68} & \textbf{76.79} \\
\bottomrule
\end{tabular}
}
\end{center}
\caption{Ablation study on LLaMA2-13B model. Results are reported as AUROC, Accuracy (Acc), and F1 on three benchmarks. The rightmost columns show averaged metrics across all datasets.}
\label{tab:ablation_main_appendix}
\end{table*}

\begin{figure*}[t]
    \centering
    
    \begin{subfigure}[t]{0.48\textwidth}
        \centering
        \includegraphics[width=\linewidth]{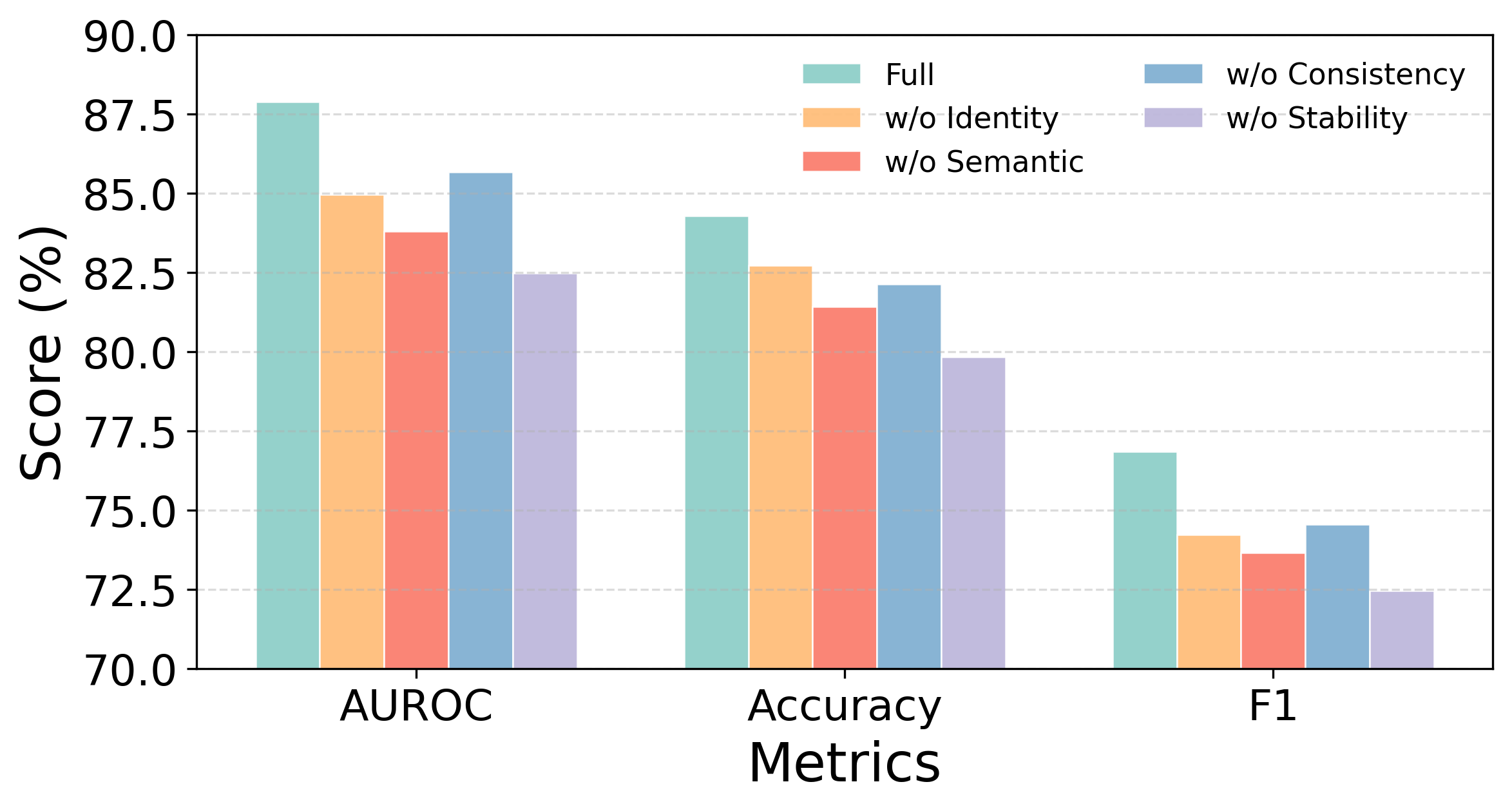}
        \caption{Ablation Result on RAGTruth}
        \label{fig:ablation_ragtruth}
    \end{subfigure}
    \hfill
    \begin{subfigure}[t]{0.48\textwidth}
        \centering
        \includegraphics[width=\linewidth]{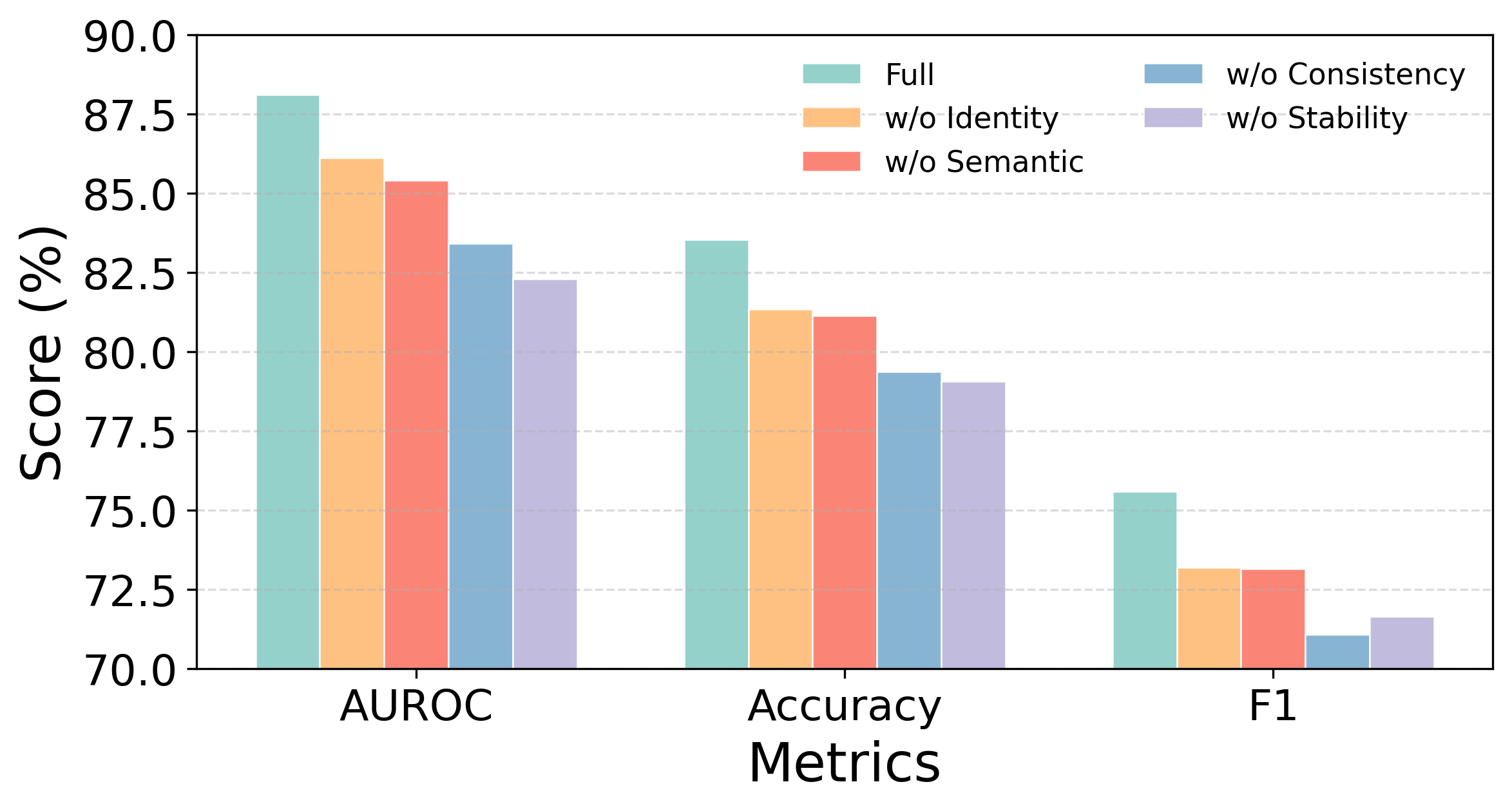}
        \caption{Ablation Result on HotpotQA}
        \label{fig:ablation_hotpotqa}
    \end{subfigure}
    

    \begin{subfigure}[t]{0.48\textwidth}
        \centering
        \includegraphics[width=\linewidth]{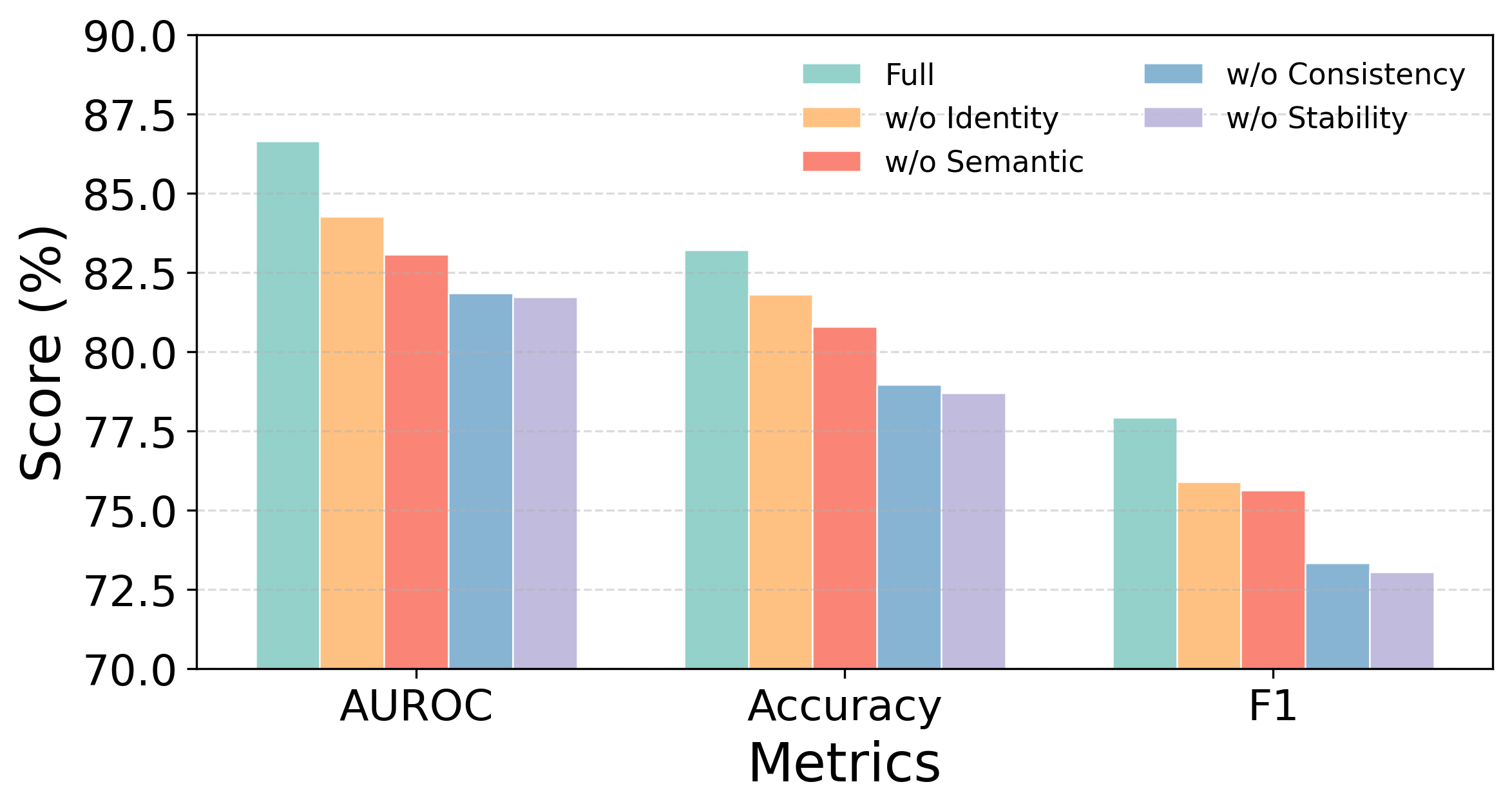}
        \caption{Ablation Result on DelucionQA}
        \label{fig:ablation_delucionqa}
    \end{subfigure}
    \hfill
    \begin{subfigure}[t]{0.48\textwidth}
        \centering
        \includegraphics[width=\linewidth]{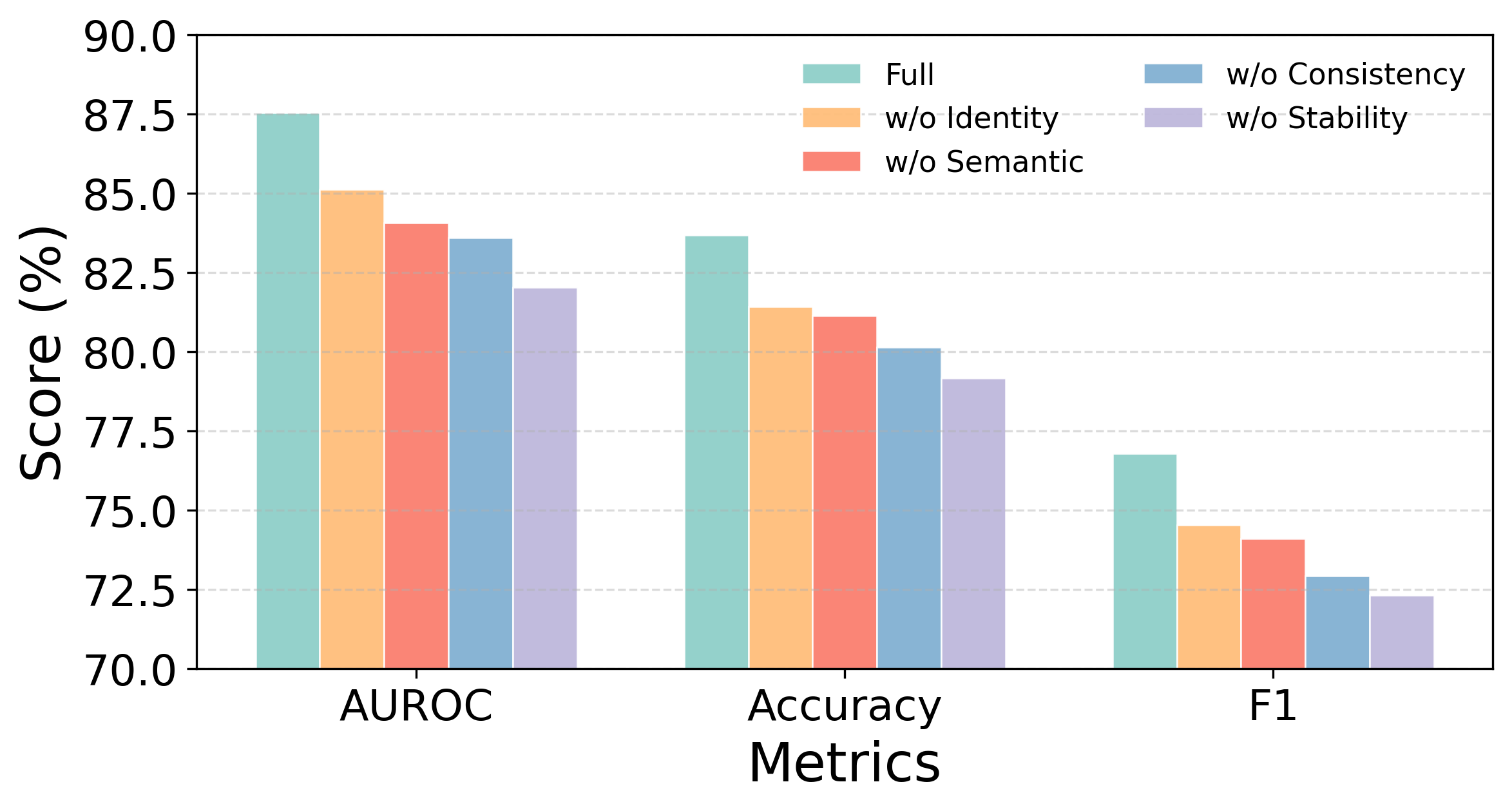}
        \caption{Average Ablation Result}
        \label{fig:ablation_average}
    \end{subfigure}
    
    \caption{Detailed ablation results on LLaMA2-13B. 
    Results are shown on RAGTruth (top-left), HotpotQA (top-right), DelucionQA (bottom-left), 
    and averaged across datasets (bottom-right).}
    \label{fig:ablation_appendix}
\end{figure*}

We provide comprehensive ablation analysis across all datasets and model architectures to validate each component's contribution. Table~\ref{tab:ablation_main_appendix} presents the complete ablation results on LLaMA2-13B, while Figures~\ref{fig:ablation_ragtruth} through \ref{fig:ablation_average} show detailed performance degradation patterns across individual datasets and averaged results.

The ablation visualizations reveal distinct component contributions across different evaluation scenarios. Counterfactual stability analysis demonstrates the most substantial impact across all datasets, with removal leading to 5.52 AUROC points average degradation. This consistent pattern confirms the necessity of distinguishing genuine evidence support from spurious correlations regardless of dataset characteristics. Consistency alignment shows particularly pronounced effects on HotpotQA and DelucionQA, where quantitative verification becomes critical for multi-hop reasoning and domain-specific content. Semantic alignment exhibits stronger influence on RAGTruth, reflecting its importance for handling paraphrased expressions in general knowledge contexts. Identity alignment provides steady baseline performance through exact matching across all evaluation settings.

The balanced degradation curves across datasets validate our multi-dimensional design philosophy. Each alignment dimension addresses distinct verification challenges while maintaining complementary effects, with no single component dominating performance. The stability analysis component's consistent importance across all scenarios confirms the practical value of robustness testing in evidence-based verification systems.

\subsubsection{Sensitivity Analysis}
\label{sec:sensitivity_details}
\begin{figure*}[t]
    \centering
    
    \begin{subfigure}[t]{0.48\textwidth}
        \centering
        \includegraphics[width=\linewidth]{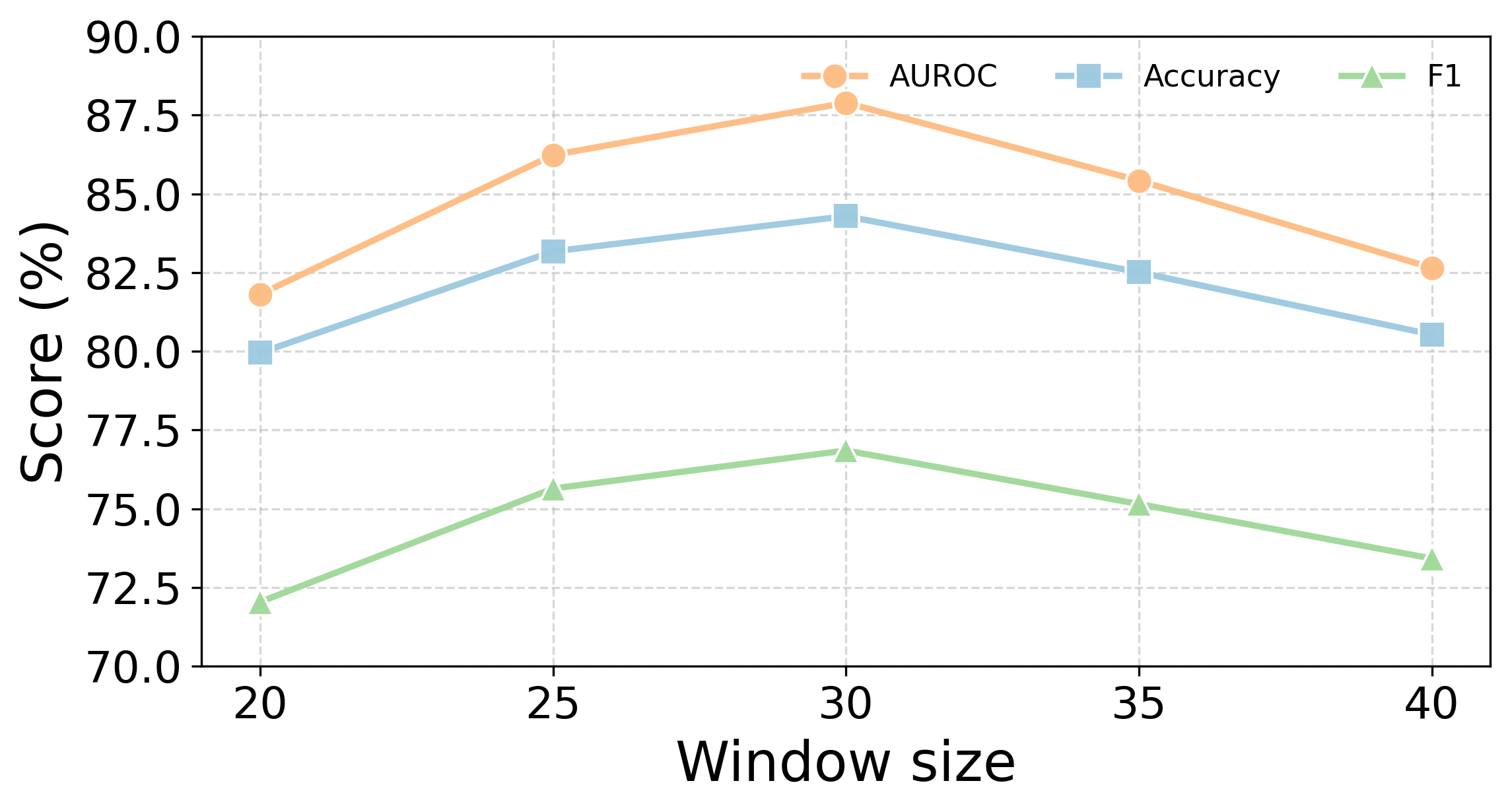}
        \caption{Window Sensitivity on RAGTruth}
        \label{fig:sensitivity_ragtruth}
    \end{subfigure}
    \hfill
    \begin{subfigure}[t]{0.48\textwidth}
        \centering
        \includegraphics[width=\linewidth]{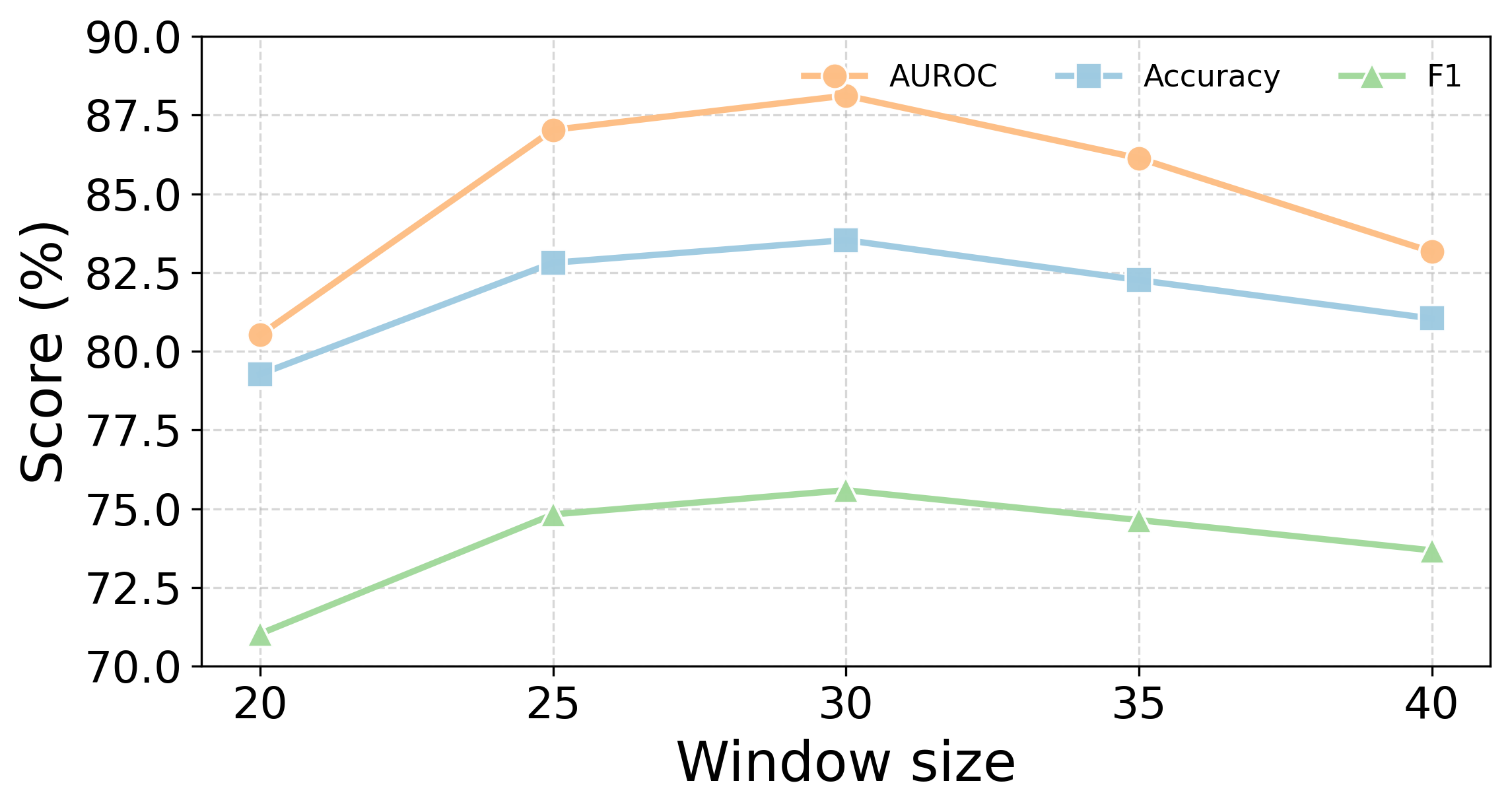}
        \caption{Window Sensitivity on HotpotQA}
        \label{fig:sensitivity_hotpotqa}
    \end{subfigure}
    
    \par\smallskip

    \begin{subfigure}[t]{0.48\textwidth}
        \centering
        \includegraphics[width=\linewidth]{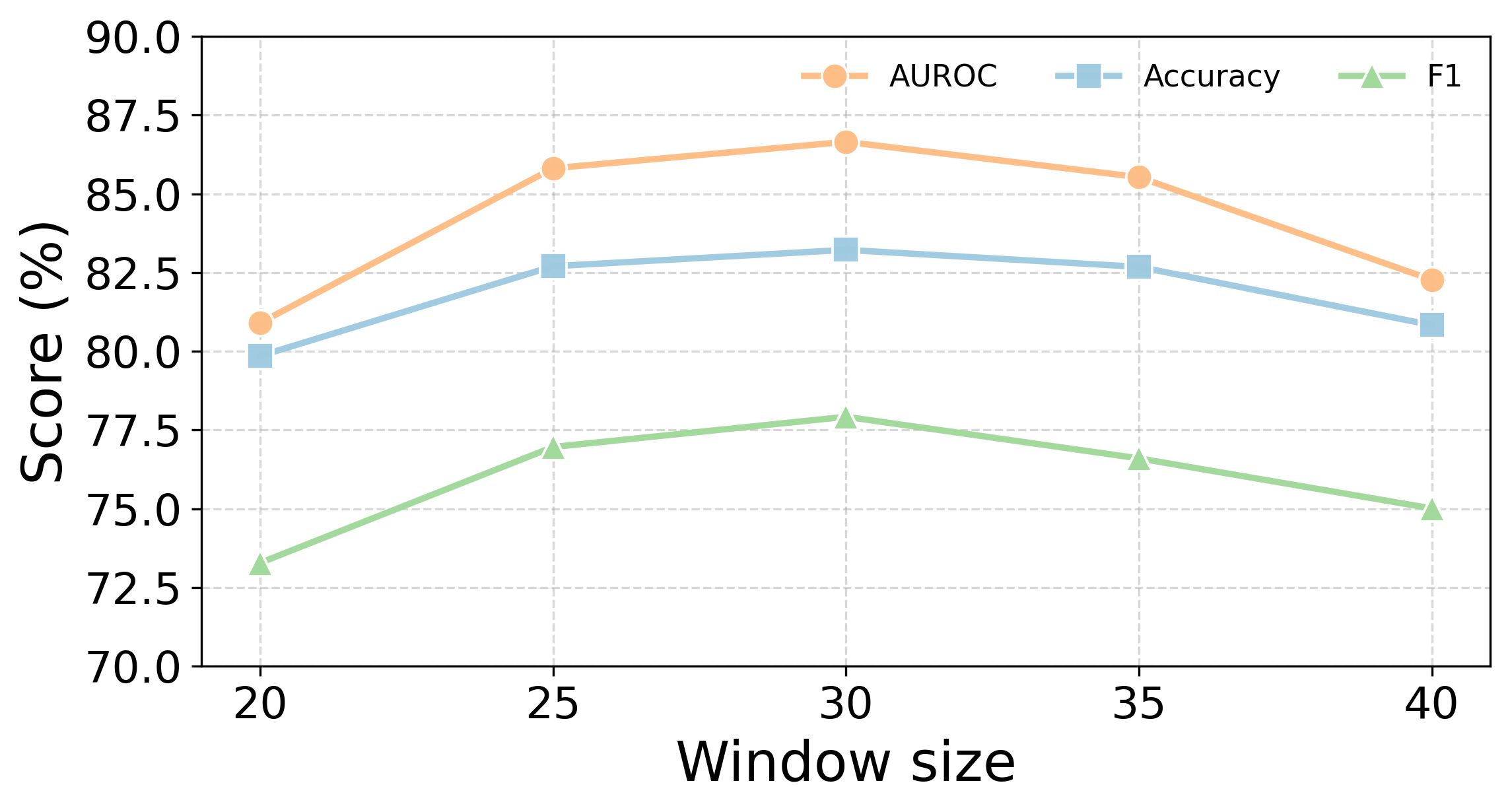}
        \caption{Window Sensitivity on DelucionQA}
        \label{fig:sensitivity_delucionqa}
    \end{subfigure}
    \hfill
    \begin{subfigure}[t]{0.48\textwidth}
        \centering
        \includegraphics[width=\linewidth]{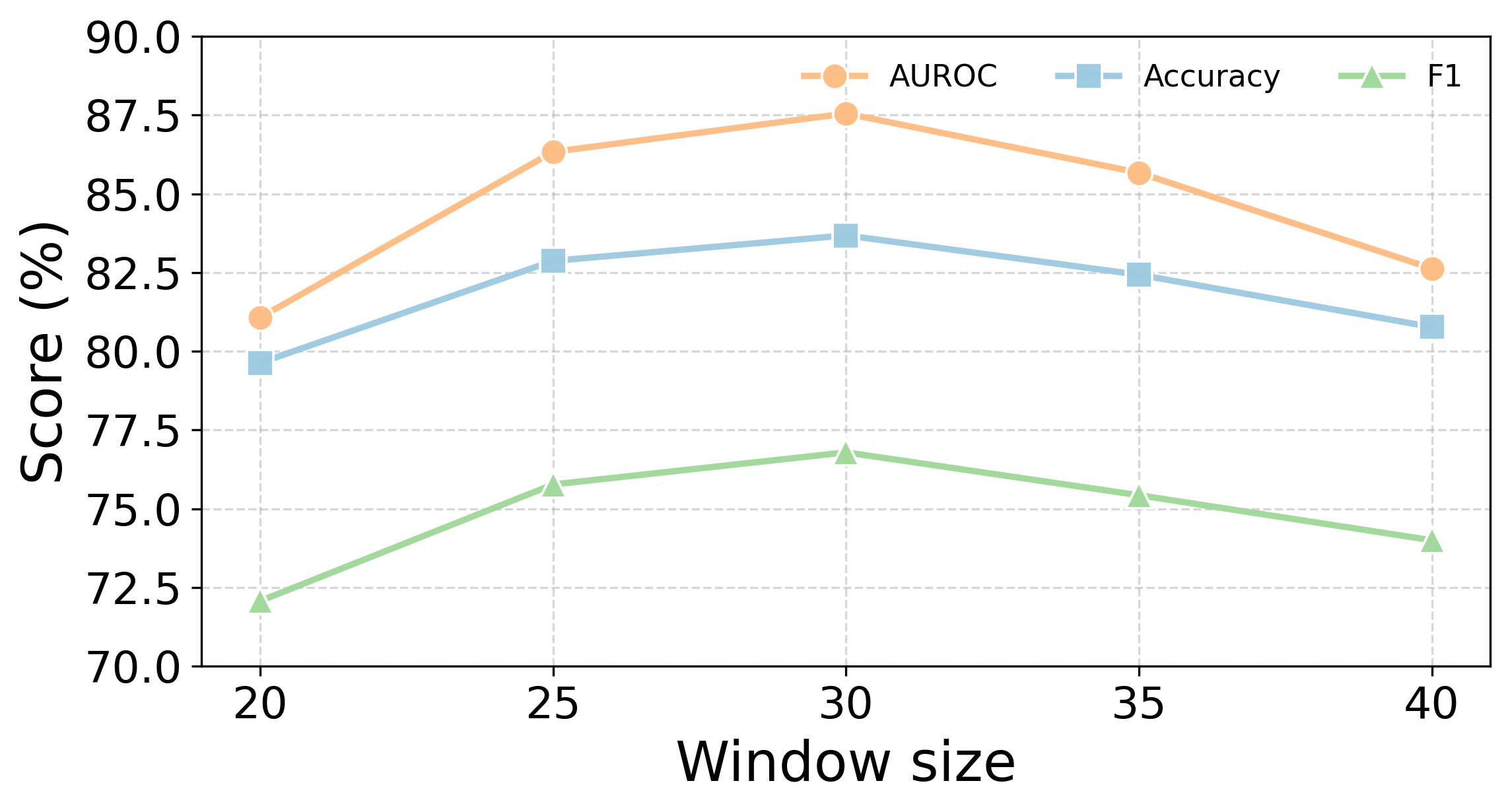}
        \caption{Average Window Sensitivity}
        \label{fig:sensitivity_average}
    \end{subfigure}
    
    \caption{Parameter sensitivity analysis of \textbf{EAEV} under different answer-side window lengths. 
    Results are shown on RAGTruth (top-left), HotpotQA (top-right), DelucionQA (bottom-left), 
    and averaged across datasets (bottom-right).}
    \label{fig:sensitivity_appendix}
\end{figure*}

Parameter sensitivity analysis demonstrates EAEV's robustness across different configuration settings. Table~\ref{tab:sensitivity_window_final} provides detailed performance under varying answer-side window lengths, while Figures~\ref{fig:sensitivity_ragtruth} through \ref{fig:sensitivity_average} illustrate the characteristic inverted-U performance curves across individual datasets.

\begin{table*}[t]

\begin{center}
\resizebox{\textwidth}{!}{
\begin{tabular}{l ccc ccc ccc ccc}
\toprule
& \multicolumn{3}{c}{\textbf{RAGTruth}}
& \multicolumn{3}{c}{\textbf{HotpotQA}}
& \multicolumn{3}{c}{\textbf{DelucionQA}}
& \multicolumn{3}{c}{\textbf{Average}} \\
\cmidrule(lr){2-4}\cmidrule(lr){5-7}\cmidrule(lr){8-10}\cmidrule(lr){11-13}
\textbf{window size}
& AUROC & Acc & F1
& AUROC & Acc & F1
& AUROC & Acc & F1
& AUROC & Acc & F1 \\
\midrule
20 & 81.81 & 79.95 & 72.04 & 80.52 & 79.27 & 71.03 & 80.90 & 79.85 & 73.28 & 81.07 & 79.63 & 72.07 \\
25 & 86.23 & 83.17 & 75.64 & 87.03 & 82.81 & 74.82 & 85.81 & 82.70 & 76.96 & 86.33 & 82.87 & 75.77 \\
\textbf{30} & \textbf{87.89} & \textbf{84.29} & \textbf{76.85}
            & \textbf{88.12} & \textbf{83.53} & \textbf{75.59}
            & \textbf{86.65} & \textbf{83.22} & \textbf{77.92}
            & \textbf{87.55} & \textbf{83.68} & \textbf{76.79} \\
35 & 85.42 & 82.51 & 75.15 & 86.13 & 82.26 & 74.64 & 85.53 & 82.68 & 76.60 & 85.67 & 82.43 & 75.43 \\
40 & 82.65 & 80.53 & 73.42 & 83.16 & 81.04 & 73.68 & 82.27 & 80.83 & 75.01 & 82.63 & 80.77 & 74.00 \\
\bottomrule
\end{tabular}
}
\end{center}
\caption{Sensitivity to answer-side window length (LLaMA2-13B). Results are reported as AUROC, Accuracy (Acc), and F1 on three benchmarks. The rightmost columns show averaged metrics across all datasets.}
\label{tab:sensitivity_window_final}
\end{table*}

The sensitivity visualizations reveal consistent optimal performance at 30-token windows across all datasets, with graceful degradation patterns for both smaller and larger window sizes. RAGTruth shows the sharpest sensitivity curve, indicating that fine-grained annotations benefit most from optimal contextualization. HotpotQA exhibits broader stability around the optimum, reflecting the method's robustness for multi-hop reasoning tasks. DelucionQA demonstrates intermediate sensitivity patterns, suggesting balanced requirements between contextual information and noise control in domain-specific settings.

Performance degradation below 25 tokens reflects insufficient contextual information for accurate alignment assessment, while degradation above 35 tokens indicates noise introduction from irrelevant content. The framework maintains reasonable stability within the 25-35 token range across all datasets, supporting practical deployment flexibility. These results validate our parameter selection methodology and confirm EAEV's reliability under varying configuration requirements.

\subsubsection{Results on a Newer Backbone LLM}
\label{sec:appendix_new_models}

To address concerns about model freshness, we additionally evaluate our approach on a newer backbone, Qwen3-8B, using the RAGTruth dataset. We run several strong baselines under the same evaluation protocol. Table~\ref{tab:qwen3_ragtruth} reports Precision, Recall, and F1. Our method achieves the best overall performance, indicating that the effectiveness of EAEV generalizes to newer LLM backbones.

\begin{table}[t]

\centering
\setlength{\tabcolsep}{4pt}  
\begin{tabular}{l l c c c}
\toprule
\textbf{Model} & \textbf{Method} & \textbf{Precision} & \textbf{Recall} & \textbf{F1} \\
\midrule
\multirow{4}{*}{Qwen3-8B} 
& RefCheck & 75.54 & 76.66 & 69.43 \\
& ReDEeP  & 78.85 & 78.98 & 72.35 \\
& TSV    & 81.56 & 80.08 & 72.23 \\
& \textbf{Ours} & \textbf{87.51} & \textbf{81.56} & \textbf{75.39} \\
\bottomrule
\end{tabular}
\caption{Performance on RAGTruth using a newer backbone model (Qwen3-8B). Our method remains consistently strong compared to representative baselines.}
\label{tab:qwen3_ragtruth}
\end{table}

\subsubsection{Entity-Level Hallucination Detection on RAGTruth}
\label{sec:appendix_entity_level}

While the main paper reports answer-level detection results, EAEV is designed to provide fine-grained, entity-anchored verification signals. We therefore additionally report entity-level hallucination detection on RAGTruth. Table~\ref{tab:entity_level_ragtruth} shows AUROC, Accuracy, and F1 for our method across three backbone models. The results demonstrate that EAEV yields meaningful entity-level separability between supported and unsupported entities, consistent with the goal of fine-grained verification.

\begin{table}[t]

\centering
\begin{tabular}{l c c c}
\toprule
\textbf{Model} & \textbf{AUROC} & \textbf{Acc} & \textbf{F1} \\
\midrule
Qwen2.5-7B   & 77.42 & 78.12 & 67.12 \\
LLaMA2-7B    & 76.95 & 75.87 & 66.58 \\
LLaMA2-13B   & 78.34 & 78.68 & 68.36 \\
\bottomrule
\end{tabular}
\caption{Entity-level hallucination detection performance of EAEV on RAGTruth across different backbones.}
\label{tab:entity_level_ragtruth}
\end{table}

\subsubsection{Resource Cost Comparison}
\label{sec:appendix_cost}

We report an efficiency comparison between EAEV and SelfCheckGPT under the same setting (50 samples; Qwen2.5-7B-Instruct). Table~\ref{tab:cost_selfcheck} summarizes runtime and memory usage. EAEV is substantially faster and more memory efficient in this setup, largely because SelfCheckGPT requires multiple additional LLM forward passes per sample and similarity computations (e.g., BERTScore), whereas EAEV relies on a single generation with evidence-aligned scoring.

\begin{table*}[t]

\centering
\begin{tabular}{l c c c}
\toprule
\textbf{Metric} & \textbf{EAEV} & \textbf{SelfCheckGPT} & \textbf{Efficiency Ratio} \\
\midrule
Total Runtime (s) & 1825.9 & 13348.5 & 7.3$\times$ faster \\
GPU Memory (MB)   & 6578.7 & 7735.4  & lower \\
CPU Memory (MB)   & 42.5   & 146.3   & lower \\
\bottomrule
\end{tabular}
\caption{Resource cost comparison between EAEV and SelfCheckGPT (50 samples, Qwen2.5-7B-Instruct).}
\label{tab:cost_selfcheck}
\end{table*}

\subsection{Case Study}

\begin{figure}[t]
  \centering
  \includegraphics[width=\columnwidth]{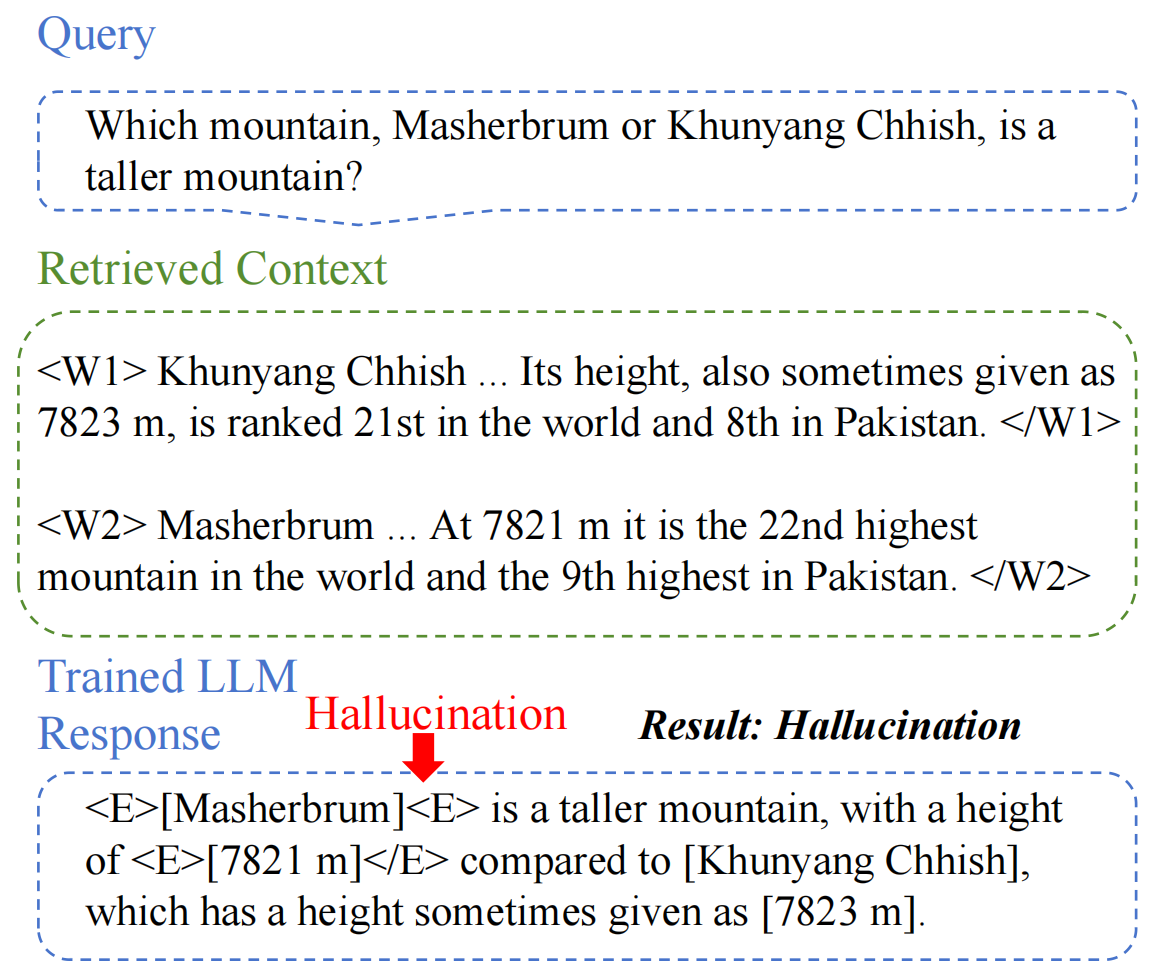}
  \caption{
  An illustrative case study of entity-level hallucination detection with EAEV.
  Although the retrieved context contains correct height information for both mountains,
  the generated answer incorrectly binds the comparative claim to an inconsistent numerical entity.
  EAEV explicitly identifies unsupported entities (\textless E\textgreater) while preserving supported ones.
  }
  \label{fig:case_study}
\end{figure}

To qualitatively illustrate how EAEV performs entity-level hallucination detection under RAG settings, we present a representative case study in Figure~\ref{fig:case_study}. 
The example demonstrates a scenario where the retrieved context contains correct factual evidence, yet the generated answer exhibits a subtle entity-level inconsistency due to incorrect binding between a comparative claim and numerical attributes. 
EAEV explicitly marks unsupported entities while preserving supported ones, enabling fine-grained and interpretable hallucination detection beyond sentence-level semantic coherence.

\subsection{Details for Prompt Template}

The EAEV-guided supervised fine-tuning transforms entity verification into an executable annotation generation task. We design a structured prompt that enables models to learn EAEV's multi-dimensional alignment patterns through standard supervised training while maintaining evidence traceability.

\begin{table}[t]

  \centering
  \begin{tabular}{p{\columnwidth}}
    \toprule
    \textbf{Task} \\
    \midrule
    Given a question, supporting passages, and a model answer, mark any unsupported or contradictory
    entity mentions in the answer using \texttt{<E>...\allowbreak</E>} tags. Keep the original answer text and only add tags. \\
    \midrule
    \textbf{Question} \\
    \textit{\{String\}} \\
    \midrule
    \textbf{Supporting Passages} \\
    \midrule
    \texttt{<W1>} \textit{\{String\}} \texttt{</W1>} \\
    \texttt{<W2>} \textit{\{String\}} \texttt{</W2>} \\
    \texttt{<W3>} \textit{\{String\}} \texttt{</W3>} \\
    \midrule
    \textbf{Original Answer} \\
    \textit{\{String\}} \\
    \midrule
    \textbf{Instructions} \\
    \midrule
    $\bullet$ Mark entities contradicted by the supporting passages. \\
    $\bullet$ Mark entities lacking sufficient evidence support. \\
    $\bullet$ Preserve all original text—only add \texttt{<E>...</E>} tags. \\
    $\bullet$ Focus on named entities, dates, numbers, and key factual claims. \\
    \midrule
    \textbf{Expected Output} \\
    \midrule
    \textit{Answer text with \texttt{<E>...\allowbreak</E>} tags around unsupported entities.} \\
    \bottomrule
  \end{tabular}
  \caption{Prompt template for EAEV-guided supervised fine-tuning.}
  \label{tab:eaev_prompt}
\end{table}

The prompt design follows several key principles to ensure effective knowledge transfer from EAEV's verification framework. The task description explicitly requires minimal modification where only annotation tags are added without altering original answer text. Supporting passages are clearly delineated with window markers (\texttt{<W1>}, \texttt{<W2>}, \texttt{<W3>}) to maintain precise evidence traceability throughout verification. The instructions distinguish between contradicted entities and those lacking sufficient support, reflecting EAEV's multi-dimensional alignment assessment.

This structured approach enables standard supervised fine-tuning to learn sophisticated verification patterns while preserving interpretability through direct evidence grounding. The prompt transforms entity-level hallucination detection into a sequence labeling task that models can learn through token-weighted cross-entropy loss, directly implementing the supervision mechanism described in Section~\ref{sec:methodology}.

\subsection{Usage Claim of AI assistants}

We use LLM for grammar and spelling checks only, with prompt ``Proofread the sentences''. All conceptual development, analysis, writing, and editing were carried out solely by the authors without LLM assistance.

\end{document}